\documentclass[sigconf]{acmart} 
\usepackage{multirow}
\usepackage[normalem]{ulem} 
\useunder{\uline}{\ul}{}
\usepackage{xspace}
\usepackage{xcolor}
\usepackage{lipsum}
\usepackage{tabularx}
\usepackage[most]{tcolorbox}
\tcbuselibrary{skins,breakable,listings}

\usepackage{fvextra}
\newsavebox{\safeboxbox}

\newenvironment{safebox}
  {\VerbatimEnvironment
   \begin{lrbox}{\safeboxbox}
   \begin{minipage}{0.97\linewidth}
   \normalfont\small\raggedright
   \begin{Verbatim}[
     breaklines=true,
     breakanywhere=true,
     breaksymbolleft={},
     breaksymbolright={}
   ]}
  {\end{Verbatim}
   \end{minipage}
   \end{lrbox}
   \noindent\fbox{\usebox{\safeboxbox}}}

\definecolor{bluegray}{RGB}{230,234,238}
\definecolor{bluegrayframe}{RGB}{150,158,168}

\definecolor{tealgray}{RGB}{230,238,236}
\definecolor{tealgrayframe}{RGB}{140,165,160}

\definecolor{violetgray}{RGB}{234,230,238}
\definecolor{violetgrayframe}{RGB}{160,145,170}

\usepackage{listings}
\newcommand{\NAME}{{GraphScout}\xspace}
\newcommand\zty[1]{{\color{black}#1}}
\AtBeginDocument{%
  }

\setcopyright{acmlicensed}
\copyrightyear{2018}
\acmYear{2018}
\acmDOI{XXXXXXX.XXXXXXX}

\acmConference[Conference acronym 'XX]{Make sure to enter the correct
  conference title from your rights confirmation email}{June 03--05,
  2018}{Woodstock, NY}
\acmISBN{978-1-4503-XXXX-X/2018/06}

\begin{document}

\title{\zty{GraphScout: Empowering Large Language Models with Intrinsic Exploration Ability for Agentic Graph Reasoning}}


\author{Yuchen Ying}
\authornote{These authors contributed equally to this work.}
\affiliation{%
  \institution{Zhejiang University}
  \city{Hangzhou}
  \country{China}}
\email{yingyc@zju.edu.cn}

\author{Weiqi Jiang}
\authornotemark[1]
\affiliation{%
  \institution{Zhejiang University}
  \city{Hangzhou}
  \country{China}}
\email{weiqi.jiang@zju.edu.cn}

\author{Tongya Zheng}
\authornote{Corresponding author.}
\affiliation{%
  \institution{Hangzhou City University}
  \city{Hangzhou}
  \country{China}}
\email{doujiang zheng@163.com}

\author{Yu Wang}
\affiliation{%
  \institution{Zhejiang University}
  \city{Hangzhou}
  \country{China}}
\email{yu.wang@zju.edu.cn}

\author{Shunyu Liu}
\affiliation{%
  \institution{Nanyang Technological University}
  \city{Singapore}
  \country{Singapore}}
\email{shunyu.liu@ntu.edu.sg }

\author{Kaixuan Chen}
\affiliation{%
  \institution{Zhejiang University}
  \city{Hangzhou}
  \country{China}}
\email{chenkx@zju.edu.cn}

\author{Mingli Song}
\affiliation{%
  \institution{Zhejiang University}
  \city{Hangzhou}
  \country{China}}
\email{brooksong@zju.edu.cn}


\begin{abstract}
 Knowledge graphs provide structured and reliable information for many real-world applications, motivating increasing interest in combining large language models (LLMs) with graph-based retrieval to improve factual grounding. Recent Graph-based Retrieval-Augmented Generation (GraphRAG) methods therefore introduce iterative interaction between LLMs and knowledge graphs to enhance reasoning capability. However, existing approaches typically depend on manually designed guidance and interact with knowledge graphs through a limited set of predefined tools, which substantially constrains graph exploration. To address these limitations, we propose GraphScout, a training-centric agentic graph reasoning framework equipped with more flexible graph exploration tools. GraphScout enables models to autonomously interact with knowledge graphs to synthesize structured training data which are then used to post-train LLMs, thereby internalizing agentic graph reasoning ability without laborious manual annotation or task curation.  Extensive experiments across five knowledge-graph domains show that a small model (e.g., Qwen3-4B) augmented with GraphScout outperforms baseline methods built on leading LLMs (e.g., Qwen-Max) by an average of 16.7\% while requiring significantly fewer inference tokens. Moreover, GraphScout exhibits robust cross-domain transfer performance.  Our code will be made publicly available~\footnote{\url{https://github.com/Ying-Yuchen/_GraphScout_}}.
\end{abstract}




\maketitle

\section{Introduction}

Large language models (LLMs)~\cite{achiam2023gpt4,touvron2023llama2,yang2025qwen3,glm2024chatglm} have achieved remarkable success in question answering and complex reasoning, but they remain prone to hallucinations and lack reliable access to domain-specific or up-to-date knowledge. Retrieval-augmented generation (RAG)~\cite{shuster2021rag1} addresses these limitations by grounding model outputs in external knowledge at inference time. While effective for unstructured text, conventional RAG pipelines struggle with structured and relational data, such as knowledge graphs, where multi-hop dependencies and structural constraints are essential. This has motivated a growing interest in graph-based retrieval and reasoning (GraphRAG)~\cite{edge2024graphrag1, gutierrez2024hipporag, luo2025hypergraphrag}, which seeks to exploit graph structure for relational inference and information aggregation.

The first class of GraphRAG methods consists of passive retrieval-driven approaches. These methods~\cite{jin202llm_on_graphs,ye2024language} are typically built upon the standard RAG framework. They first employ static retrieval strategies to select a set of nodes from the knowledge graph that are relevant to the given query, and then expand their neighboring subgraphs according to predefined rules, such as a fixed hop count. The resulting structured information is subsequently linearized into textual form and provided to LLMs for downstream reasoning.

The second class comprises active traversal-based approaches. These methods~\cite{sun2023tog, jin2024graphcot, luo2025graphr1, yu2025graphrag} equip LLMs with a set of basic graph interaction tools, such as node querying and relation expansion, and leverage carefully designed prompting schemes to enable multi-round interactions with the knowledge graph during the reasoning process. Active traversal-based approaches enable dynamic graph exploration, enhancing retrieval–reasoning coordination. Recent methods~\cite{kashmira2025graphrunner, gao2025graphCounselor, liu2025polyg, liu2025Graph-O1} further structure this process through explicit traversal planning, schema-aware validation, or multi-agent coordination, improving robustness and accuracy for complex graph reasoning.

\begin{figure}[]
    \centering    \includegraphics[width=0.93\columnwidth]{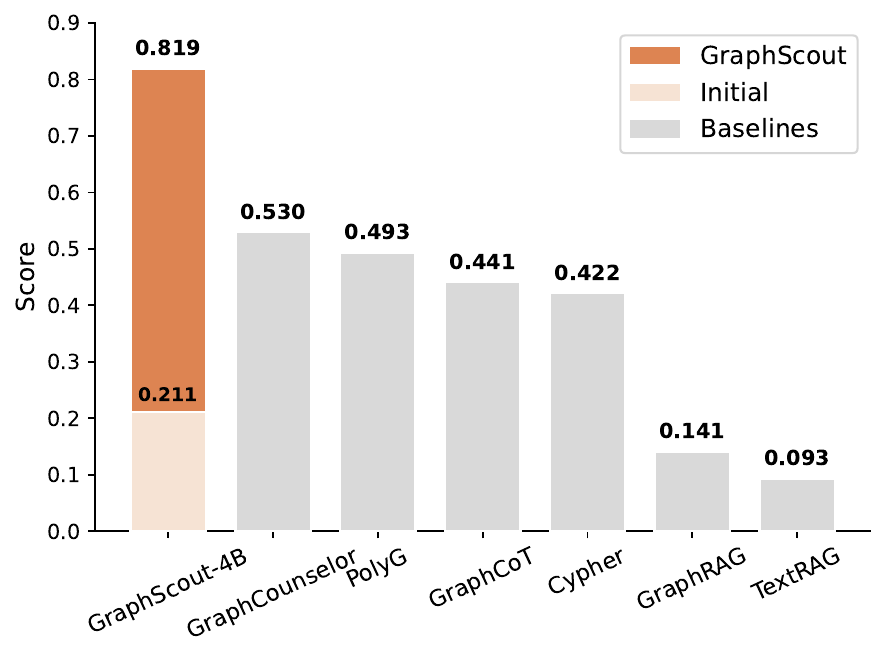}
 \caption{\zty{Qwen3-4B-Instruct with \NAME achieves substantial gains over leading LLMs (e.g.,Qwen-Max) with prompting-based GraphRAG baselines in Healthcare dataset.}}
    \label{fig:head_pic}
\end{figure}

Despite recent progress in GraphRAG, existing methods still suffer from notable limitations in flexibility and graph exploration. On the one hand, most current approaches rely on manually designed and limited sets of graph interaction primitives, which constrains their expressiveness and flexibility. For example, when performing multi-hop neighbor queries along known path types, models often have to invoke basic neighbor expansion operations iteratively, resulting in prolonged interaction sequences and low efficiency. On the other hand, LLMs lack intrinsic priors for structured graph exploration, causing these methods to primarily depend on external mechanisms, such as workflow constraints, to enhance graph exploration. As a result, they struggle to achieve a fundamental improvement in the LLM’s intrinsic graph reasoning and exploration capabilities. Empirical results reported in prior work (e.g., GraphCoT~\cite{jin2024graphcot} in Figure~\ref{fig:head_pic}) also reflect these limitations to some extent. Even with more flexible graph exploration tools, these issues remain unresolved. For example, PolyG~\cite{liu2025polyg} adopts a Cypher-based interface to enable more general graph operations, yet its performance (in Figure~\ref{fig:head_pic}) only marginally surpasses that of GraphCoT due to its limited intrinsic graph reasoning and exploration abilities.

These observations suggest that current GraphRAG research calls for a shift in methodological perspective. Rather than continually stacking increasingly complex prompt engineering or workflow control mechanisms, a more promising direction is to equip LLMs with more flexible and expressive graph interaction tools, while fundamentally enhancing their intrinsic graph reasoning and exploration abilities. Nevertheless, enhancing such intrinsic capabilities typically requires targeted post-training, which in turn relies on diverse, high-quality annotated graph reasoning trajectories that are expensive to manually construct and difficult to scale in practice, forming a major bottleneck.

To address the above challenges, we propose \textbf{GraphScout}, a training-centric framework designed to equip LLM with intrinsic exploration ability for agentic graph reasoning. GraphScout consists of three core components. First, GraphScout provides a set of \textbf{Agentic Graph Exploration Tools}, including a Code Interpreter and a Node Retriever, which enable LLMs to interact with the knowledge graph in a highly flexible and programmable manner. Second, the \textbf{Graph Quizzer} module leverages a strong LLM as a senior scout to freely explore the graph environment, automatically synthesizing diverse and high-quality graph query–answer pairs, along with their corresponding evidence clues that serve as supervision signals for subsequent training. Finally, the \textbf{Graph Solver} module optimizes small-parameter LLM as a junior scout under a multi-turn interaction paradigm, enabling it to progressively acquire agentic graph reasoning ability.
Experiments show that LLMs trained with GraphScout consistently outperform baselines built on substantially larger backbone LLMs on multiple GraphRAG benchmarks by an average margin of \textbf{16.7\%}, while requiring an order of magnitude fewer inference tokens. Moreover, models trained on a single-domain knowledge graph transfer robustly to unseen domains, demonstrating stable cross-domain performance.

In summary, our contributions are as follows:
\begin{itemize}

    \item We identify the flexibility and graph exploration challenges of existing GraphRAG methods, which largely rely on manually designed interaction tools and external control mechanisms, and highlight the need for enabling LLMs with intrinsic
exploration ability for agentic graph reasoning.

 \item We propose \textbf{GraphScout}, a training-centric framework that equips LLMs with Agentic Graph Exploration Tools for intrinsic graph reasoning. 
GraphScout integrates Graph Quizzer to autonomously construct diverse, high-quality graph question–answer pairs, and Graph Solver to post-train LLMs to internalize effective graph exploration ability.
    
    \item Extensive experiments across multiple GraphRAG datasets demonstrate that GraphScout enables small-parameter LLMs to achieve strong performance and efficiency, surpassing baseline methods built on flagship LLMs by an average margin of 16.7\%, while exhibiting robust cross-domain generalization and reduced inference cost.
\end{itemize}

\begin{figure*}[h]
    \centering 
    \includegraphics[width=0.95\linewidth]{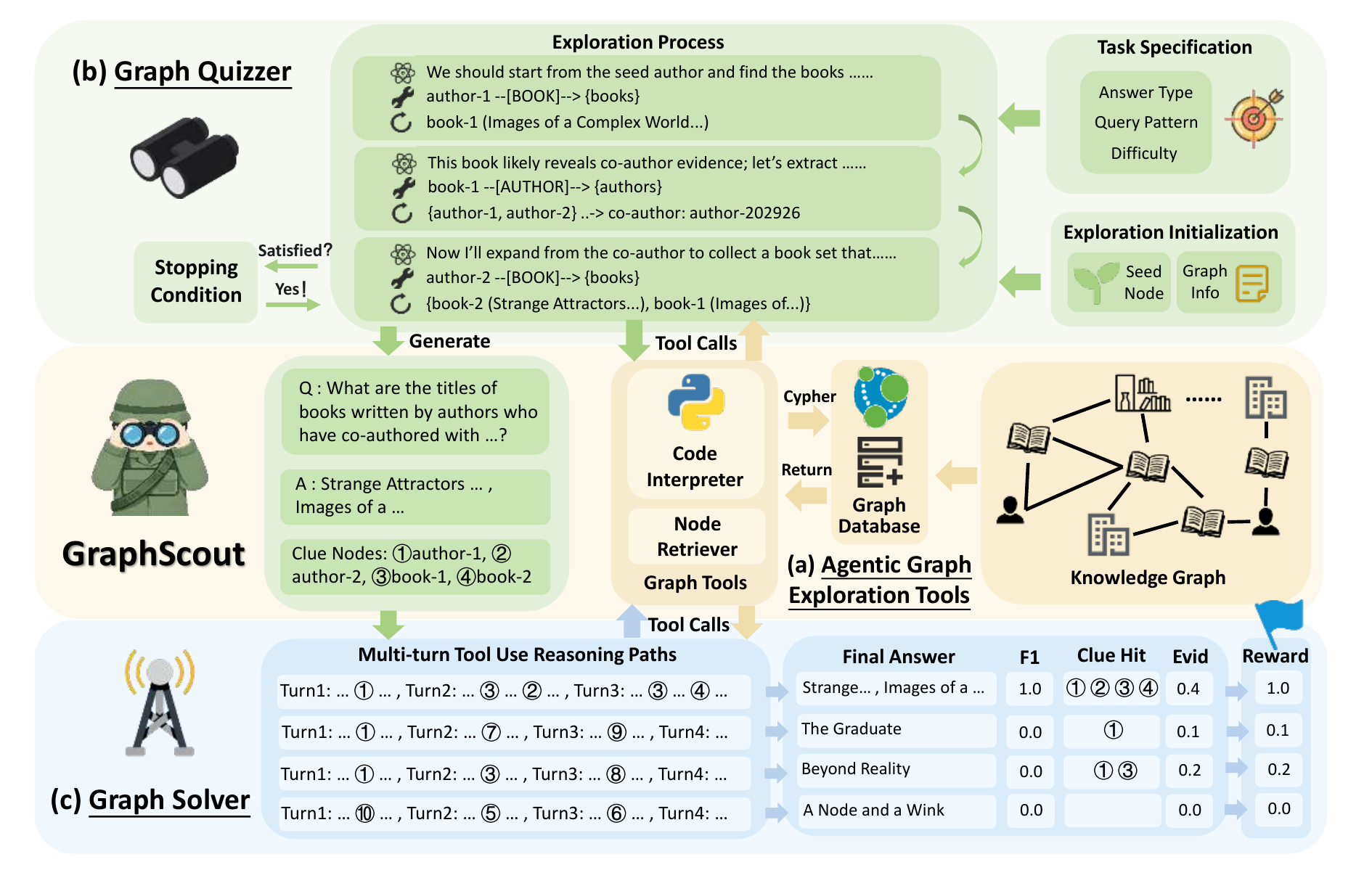}
 \caption{Overview of \NAME framework.}
    \label{fig:graphscout}
\end{figure*}

\section{Related Work}
\subsection{LLM for Graph Reasoning}
Inspired by the recent success of large language models (LLMs) in natural language processing, increasing attention has been paid to applying LLMs to graph-related tasks~\cite{jin202llm_on_graphs}.
Existing studies can be broadly categorized according to the role LLMs play in graph reasoning. One line of work treats LLMs as feature extractors~\cite{chen2024llm_extractor1,chien2021llm_extractor2,li2025llm_extractor3,duan2023llm_extractor4,zhao2022llm_extractor5}, where textual information associated with graph nodes or edges is encoded by LLMs and then fed into graph neural networks (GNNs)~\cite{kipf2016gcn,velivckovic2017gat,xu2018gin}. These methods typically adopt an LLM-GNN cascaded architecture, leveraging LLMs for semantic representation and GNNs for structural modeling. Another line of research employs LLMs as final predictors~\cite{jin2023llm_predictor1}.
In this paradigm, graph structures are either serialized into textual sequences and processed directly by LLMs~\cite{wang2023can, chen2024graphwiz}, or combined with GNNs that are specifically trained or fine-tuned to provide structural supervision for LLM-based reasoning~\cite{chai2025graphllm,tang2024graphgpt,chen2024llaga}. Such approaches have been applied to traditional graph tasks such as node classification and link prediction~\cite{tang2024graphgpt, lin2024langgfm}, as well as graph algorithmic reasoning that requires structural understanding~\cite{fatemi2023talk, luo2024graphinstruct,dai2024large,huang2025graphthought,guo2025g1}. In contrast to the above approaches, our work focuses on enabling LLMs to perform explicit reasoning over knowledge graphs, rather than using graphs solely as feature sources or implicit structural cues.
\subsection{Augmenting LLMs with Knowledge Graph}
Despite their strong reasoning capabilities~\cite{touvron2023llama2,achiam2023gpt4, yang2025qwen3}, LLMs suffer from limitations such as hallucinations~\cite{rawte2023hallucination2,tonmoy2024hallucination1} and outdated knowledge, motivating retrieval-augmented generation (RAG)~\cite{karpukhin2020rag2,gao2023rag3,chen2024rag4}. While effective for unstructured text, conventional RAG struggles with relational knowledge~\cite{kunegis2013konect, lastfm}.
To address this gap, recent work augments LLMs with knowledge graphs, giving rise to GraphRAG~\cite{edge2024graphrag1}. Existing approaches can be broadly categorized into two categories. The first follows a passive retrieval-driven paradigm, retrieving relevant nodes or subgraphs and linearizing them into textual representations for LLM consumption~\cite{edge2024graphrag1, guo2024lightrag, chen2024graphwiz}. 
The second line of work adopts an active traversal-based paradigm, enabling LLMs to iteratively interact with knowledge graphs by selecting traversal operations step by step~\cite{sun2023tog, jin2024graphcot}.
Representative methods include GraphCoT~\cite{jin2024graphcot}, which aligns chain-of-thought reasoning with stepwise graph traversal.
More recent approaches further extend this paradigm through structured planning or agent collaboration~\cite{kashmira2025graphrunner, gao2025graphCounselor, liu2025Graph-O1, liu2025polyg}, improving robustness and accuracy in complex graph reasoning.
PolyG~\cite{liu2025polyg} generates Cypher queries conditioned on the question structure, while GraphCounselor~\cite{gao2025graphCounselor} leverages multi-agent coordination for planning and verification.
Despite these advances, such methods still rely on manually designed interaction schemes and external control mechanisms, rather than learning intrinsic graph reasoning and exploration capabilities.

\section{GraphScout}
\zty{As illustrated in Figure~\ref{fig:graphscout}, \NAME overcomes the limitations of predefined tools by integrating a code interpreter and a node retriever as Agentic Graph Exploration Tools for both Graph Quizzer and Graph Solver.
Graph Quizzer is designed to explore high-quality graph question-answer pairs for LLM post-training, employing a strong LLM as a senior scout to construct diverse graph problems along with their corresponding traversal clues.
Graph Solver is proposed to endow LLMs with intrinsic exploration by post-training a small-parameter LLM as a junior scout, formulating graph reasoning as learning a multi-turn decision policy for agentic graph interaction. Training is primarily driven by answer rewards, with auxiliary clue-based rewards to guide exploration under sparse feedback.
Consequently, the overall GraphScout framework comprehensively explores diverse graph reasoning problems through the Graph Quizzer and efficiently addresses intrinsic graph exploration challenges via the Graph Solver.
}

\subsection{Preliminary}
\zty{Formally, a knowledge graph is denoted by $\mathcal{K} \subseteq \mathcal{E} \times \mathcal{R} \times \mathcal{E}$, where $\mathcal{E}$ is the set of entities and $\mathcal{R}$ is the set of relations.
A fact in a knowledge graph is commonly represented as a triple $(h, r, t)$ for convenience, where $h, t \in \mathcal{E}$.
The graph-based question answering task under a retrieval-augmented setting is defined as
\begin{equation}
    f: \mathcal{Q} \times \mathcal{K} \rightarrow \mathcal{Y},
    \label{eq:qa-task}
\end{equation}
where $\mathcal{Q}$ is the set of graph reasoning questions in natural language~\cite{jin2024graphcot} and $\mathcal{Y}$ is the set of answers.
The graph reasoning process of an LLM can be modeled as a policy 
$\pi(a_t \mid q, c_t)$,
which selects the next action $a_t$ given the input question $q$ and the historical reasoning trajectory $c_t$. 
Here, $\pi(\cdot)$ is the policy distribution of an LLM, $q \in \mathcal{Q}$ is a question, $a_t \in \mathcal{A}$ is an action selection at step $t$, $c_t = (o_1, a_1, \cdots, o_{t-1}, a_{t-1}, o_t)$ is the contextual observation-action sequence.
When the process reaches an answer, the reasoning trajectory can be denoted as $\tau = (o_1, a_1, \cdots, a_T, o_T)$, where $T$ is the number of reasoning steps and $o_1 = q$.
Existing prompting-based methods~\cite{kashmira2025graphrunner,gao2025graphCounselor,liu2025polyg,liu2025Graph-O1} rely on external guidance and primarily focus on enriching the action space $\mathcal{A}$ with more hand-crafted tools and auxiliary prompts.
However, a predefined set of tools $\mathcal{A}$ and a fixed policy $\pi(\cdot)$ significantly limit the exploration potential of LLMs.
Therefore, our goal is to unlock the intrinsic exploration capability of LLMs by incorporating flexible exploration tools $\mathcal{A}_\mathrm{graph}$ and learning an optimized policy $\pi_\theta (\cdot)$ for graph reasoning.
}

\subsection{\zty{Agentic Graph Exploration Tools}}
\zty{As the adage states, ``If a craftsman wants to do his work well, he must first sharpen his tools''.
Existing GraphRAG methods typically rely on a predefined set of tools, such as retrieving neighboring nodes, checking relations, or extracting node attributes.
While simple, these approaches struggle with neighborhood explosion and multi-hop collaborative patterns.
In contrast, \NAME exposes an agentic graph exploration interface $\mathcal{A}_\mathrm{graph}$, which integrates two complementary tools: a \textbf{Code Interpreter} for high-order pattern analysis and a \textbf{Node Retriever} for global graph retrieval.
This agentic interface is invoked proactively by the LLM, formatted as 
}
\begin{lstlisting}
    <tool_call> {
        "name":"tool_name",
        "arguments":{"tool_arguments"}}
    </tool_call>.
\end{lstlisting}

\subsubsection{\textbf{Code Interpreter}}
The Code Interpreter provides a safe Python execution environment with access to a Cypher-based graph query interface, allowing the model to write executable code for precise and compositional graph queries. Through this interface, the model can flexibly control query logic, process intermediate results, and adapt its exploration strategy based on the evolving interaction context, rather than following predefined traversal templates. In our implementation, large-scale knowledge graphs are stored in a Neo4j database. During a preprocessing stage, each dataset is imported into Neo4j and equipped with appropriate schema-level and property-level indexes, enabling efficient execution of Cypher queries.

\subsubsection{\textbf{Node Retriever}}
The Node Retriever is implemented as a FAISS-based vector search module that supports fuzzy entity grounding. Given textual mentions in questions or intermediate hypotheses, it retrieves candidate node identifiers based on semantic similarity, enabling robust mapping from language to graph entities prior to structured querying.

\zty{By combining a programmable Code Interpreter with a lightweight Node Retriever, \NAME extends the action space as $\mathcal{A}=\mathcal{A}_\mathrm{graph}\cup \mathcal{L}$, where $\mathcal{L}$ is the intrinsic verbal reasoning actions of an LLM.
Therefore, \NAME significantly expands the exploration space available to the LLM while keeping the tool interface minimal.
}
More details are provided in Appendix~\ref{app:tools}.

\subsection{\zty{Graph Quizzer}}
\label{sec:recon}
\zty{Using these agentic graph exploration tools $\mathcal{A}_\mathrm{graph}$, Graph Quizzer employs a strong LLM as a \emph{senior scout} to flexibly explore graph problems $\mathcal{Q}_\mathrm{train}$ at varying difficulty levels according to general task specifications.
As shown in the upper panel of Figure~\ref{fig:graphscout}, Graph Quizzer explores the higher-order neighborhoods and global graph semantics by alternating verbal reasoning and tool use until a stopping condition is met.
Through this exploration loop, Graph Quizzer generates high-quality graph question-answering pairs $\mathcal{D}_\mathrm{train} = (\mathcal{Q}_\mathrm{train}, \mathcal{Y}_\mathrm{train}, \mathcal{C}_\mathrm{train})$ for subsequent LLM post-training, effectively eliminating laborious dataset curation.
We explicitly record intermediate node clues $\mathcal{C}_\mathrm{train} = \{c | c=(e_i, e_j, \cdots, e_k) \text{ and } e_i \in \mathcal{E}\}$ to mitigate reward sparsity during the ensuing training process.
}

\subsubsection{\textbf{Task specification}}
\label{sec:task_spec}

Each quizzer episode begins by sampling an \emph{exploration objective} that specifies the structural and semantic requirements of the graph question to be generated. Rather than relying on using fixed templates, objectives are specified at an abstract level to guide task-oriented exploration while preserving flexibility. Each objective is characterized by a combination of \textbf{answer type}, \textbf{query pattern}, and \textbf{difficulty}, which together ensure diversity among generated questions.

\textbf{Answer type.}
The answer type defines the expected form of the final answer and the corresponding supervision signal. We consider four common types: \texttt{entity} (identifying a graph entity), \texttt{boolean} (judging whether a given statement is true or false), \texttt{number} (counting or aggregation), and \texttt{set} (a collection of entities). This diversity encourages the quizzer to produce supervision beyond simple entity retrieval.

\textbf{Query pattern.}
The query pattern specifies the abstract structural form of a graph query, indicating which components are given and which must be discovered through exploration. Following PolyG~\cite{liu2025polyg}, we adopt standard patterns including $\langle h,\_,\_ \rangle$, $\langle h,r,* \rangle$, $\langle h,*,t \rangle$, and $\langle h,r,t \rangle$, as well as hybrid patterns that combine multiple forms to support more compositional reasoning.

\textbf{Difficulty.}
The difficulty level controls the required reasoning depth and is categorized as \texttt{simple}, \texttt{medium}, or \texttt{hard}, corresponding to single-hop to deeper multi-hop reasoning. Varying difficulty allows quizzer to generate exploration trajectories with different lengths and structural complexity.

Together, these three factors define a structured yet flexible objective space that enables Graph Quizzer to cover diverse graph reasoning behaviors without relying on rigid, hand-crafted task templates. Additional details are provided in Appendix~\ref{app:recon_prompt}.

\subsubsection{\textbf{Exploration initialization}}
\label{sec:explore_init}

Given a sampled exploration objective, Graph Quizzer initializes the scout with a compact \emph{environment context} $c_\mathrm{env}$ and a concrete \emph{seed node} $e_i$, providing the minimal context required for task-oriented exploration.

\textbf{Graph description.}
The scout is provided with a concise description of the target knowledge graph, including core schema-level information such as node types, edge types, and commonly used properties. This description also specifies the underlying exploration tools $\mathcal{A}_\mathrm{graph}$, grounding the scout’s subsequent interactions in a concrete execution environment. 

\textbf{Seed node.}
Each exploration episode is anchored at a \emph{seed node} $e_i \in \mathcal{E}$ sampled from the target graph. To mitigate bias from imbalanced node-type distributions, we adopt a two-stage sampling strategy: first uniformly sampling a node type, followed by random selection of a node instance within that type. This strategy ensuring balanced coverage across graph structures and reasoning scenarios.

\subsubsection{\textbf{Exploration process}}
\label{sec:explore_process}
Starting from the initialized seed node $e_i$, the scout performs a bounded, multi-step exploration process conditioned on the sampled exploration objective. Exploration proceeds as an iterative \emph{explore-assess} loop, in which the scout actively retrieves graph evidence and continuously evaluates its progress toward satisfying the objective.

At each step, the scout formulates a concrete retrieval action, such as inspecting neighboring relations, expanding along selected paths, or aggregating graph statistics, and executes it through the graph interaction interface. The returned results are summarized into a local exploration state, which is used to assess whether the accumulated evidence is sufficient to meet the specified answer type, query pattern, and difficulty constraints.

This process is explicitly task-oriented: rather than exhaustively traversing the graph $\mathcal{K}$, the senior scout prioritizes \zty{a partial reasoning trajectory $c_t$} that is most relevant to the current objective and terminates exploration early once sufficient supporting information has been collected. To prevent unbounded interaction, the exploration length is capped by a fixed budget of tool calls (we use $T{=}10$), encouraging efficient and focused exploration behaviors. 

\subsubsection{\textbf{Question reporting}}
\label{sec:question_report}
Once the senior scout determines that the exploration objective has been satisfied, it transitions from the question-exploration stage to the question-reporting stage.
Using the collected reasoning trajectory $c_T$, the senior scout constructs a natural-language graph question $q$ together with its corresponding answer $y$, ensuring that $y$ is directly verifiable on the underlying graph.
In addition to the final question-answer pair $(q, y)$, Graph Quizzer records a set of \emph{evidence clues} $c = (e_i, e_j, \cdots, e_k)$, each element being a node identifier explicitly used during question and answer formation. 
These clues provide an explicit link between the generated answer $y$ and concrete graph entities $e_i$ encountered during exploration.
As a result, each quizzer episode generates a structured supervision tuple:
(\texttt{question},\ \texttt{answer},\ \texttt{clue\_nodes}),
where the question $q$ is expressed in natural language, the answer $y$ is grounded in the graph, and the evidence clues $c$ anchor supervision to explicit graph-level facts. 
This design enables downstream models to learn not only to produce correct answers, but also to align their reasoning processes with relevant evidence clues.

\subsection{\zty{Graph Solver}}
\label{sec:navi}
\zty{Graph Solver further trains a small-parameter LLM as a \emph{junior scout} to tackle the graph-reasoning questions $\mathcal{Q}_\mathrm{train}$ proposed by Graph Quizzer, using reward signals for answer correctness and alignment with evidence clues.
As shown in the bottom panel of Figure~\ref{fig:graphscout}, the junior scout iteratively reasons the possible solution paths via exploration tools $\mathcal{A}_\mathrm{graph}$ in a multi-turn paradigm until it outputs a final answer.
The reasoning policy can be denoted by $\pi_\theta(\tau | q)$, where $\pi_\theta(\cdot)$ is the parameterized LLM and $\tau$ is the trajectory involving verbal reasoning steps and tool actions.
This simple yet effective design empowers LLMs to perform intrinsic exploration over  questions $\mathcal{Q}_\mathrm{train}$ and complex graphs $\mathcal{K}$, going beyond the limited tools and static prompts used in existing GraphRAG methods.
}

\subsubsection{\textbf{Problem formulation}}
Graph Solver formulates graph question answering as learning a \emph{multi-turn decision policy} for task-oriented interaction with a knowledge graph. Given a natural language question $q$, the model must decide how to sequentially invoke tools $\mathcal{A}_\mathrm{graph}$ to retrieve evidence, aggregate information, and ultimately produce an answer.
\zty{Formally, each episode corresponds to a trajectory 
$\tau = (o_1, a_1, o_2, a_2, \ldots, o_T, a_T)$ of alternating observations and actions, where the initial observation is the question itself, i.e., $o_1 = q$. Actions $\{a_t\}_{t=1}^{T}$ are expressed as tool calls, and observations $\{o_t\}_{t=1}^{T}$ include structured tool responses and the model's intermediate verbal reasoning.}
The objective is not to follow predefined traversal rules, but to acquire a generalizable solver strategy $\pi_\theta(\tau \mid q)$ that coordinates entity grounding, graph traversal, and answer synthesis under a fixed interaction budget.

\begin{table*}[t]
\caption{Performance Comparison on GRBENCH Across Baseline Methods and GraphScout. Relative improvement is measured against the strongest baseline on each dataset.
On QwenScore, GraphScout achieves a 16.7\% average gain.}
\label{table:main}
\small
\setlength{\tabcolsep}{3.5pt}
\renewcommand{\arraystretch}{0.9}
    \centering
\begin{tabular}{cccccccccccc}
\toprule
 & Dataset & \multicolumn{2}{c}{Healthcare} & \multicolumn{2}{c}{Literature} & \multicolumn{2}{c}{Academic} & \multicolumn{2}{c}{E-Commerce} & \multicolumn{2}{c}{Legal} \\ \cline{2-12} 
\multirow{-2}{*}{Method} & LLM & QwenScore & F1 & QwenScore & F1 & QwenScore & F1 & QwenScore & F1 & QwenScore & F1 \\ \hline
 & GPT-4o & 0.137 & 0.048 & 0.221 & 0.064 & 0.097 & 0.080 & 0.110 & 0.080 & 0.244 & 0.110 \\
 & GLM-4.6 & 0.107 & 0.099 & 0.179 & 0.102 & 0.095 & 0.110 & 0.060 & 0.111 & 0.222 & 0.243 \\
 & Qwen-Max & 0.074 & 0.065 & 0.158 & 0.098 & 0.112 & 0.128 & 0.060 & 0.105 & 0.244 & 0.298 \\
\multirow{-4}{*}{\textbf{BaseLLM}} & DeepSeek-Chat & 0.104 & 0.075 & 0.192 & 0.111 & 0.134 & 0.140 & 0.095 & 0.115 & 0.222 & 0.265 \\ \hline
 & GPT-4o & 0.074 & 0.059 & 0.179 & 0.116 & 0.098 & 0.090 & 0.200 & 0.181 & 0.256 & 0.232 \\
 & GLM-4.6 & 0.093 & 0.085 & 0.146 & 0.091 & 0.092 & 0.105 & 0.250 & 0.241 & 0.217 & 0.253 \\

 & Qwen-Max & 0.067 & 0.085 & 0.142 & 0.092 & 0.092 & 0.106 & 0.090 & 0.148 & 0.344 & 0.406 \\
\multirow{-4}{*}{\textbf{TextRAG}} & DeepSeek-Chat & 0.085 & 0.060 & 0.196 & 0.137 & 0.146 & 0.158 & 0.115 & 0.138 & 0.333 & 0.365 \\ \hline
 & GPT-4o & 0.156 & 0.129 & 0.217 & 0.136 & 0.105 & 0.092 & 0.315 & 0.308 & 0.239 & 0.233 \\
 & GLM-4.6 & 0.156 & 0.174 & 0.213 & 0.164 & 0.124 & 0.135 & 0.325 & 0.350 & 0.172 & 0.253 \\

 & Qwen-Max & 0.141 & 0.168 & 0.208 & 0.166 & 0.114 & 0.130 & 0.180 & 0.243 & 0.356 & 0.424 \\
\multirow{-4}{*}{\textbf{GraphRAG}} & DeepSeek-Chat & 0.170 & 0.140 & 0.221 & 0.151 & 0.177 & 0.164 & 0.165 & 0.167 & 0.361 & 0.381 \\ \hline
 & GPT-4o & 0.311 & 0.107 & 0.533 & 0.157 & 0.605 & 0.197 & 0.485 & 0.101 & 0.389 & 0.118 \\
 & GLM-4.6 & 0.319 & 0.104 & 0.583 & 0.145 & 0.538 & 0.147 & 0.490 & 0.114 & 0.422 & 0.106 \\

 & Qwen-Max & 0.407 & 0.131 & 0.558 & 0.143 & 0.572 & 0.181 & 0.520 & 0.117 & 0.406 & 0.132 \\
\multirow{-4}{*}{\textbf{Cypher}} & DeepSeek-Chat & 0.422 & 0.097 & 0.583 & 0.144 & 0.580 & 0.186 & {\ul 0.545} & 0.114 & 0.350 & 0.094 \\ \hline
 & GPT-4o & 0.415 & 0.447 & 0.463 & 0.395 & 0.587 & 0.547 & 0.410 & 0.360 & 0.494 & 0.403 \\
 & GLM-4.6 & 0.396 & 0.475 & 0.504 & 0.486 & 0.629 & 0.614 & 0.380 & 0.388 & 0.511 & 0.517 \\

 & Qwen-Max & 0.415 & 0.470 & 0.479 & 0.437 & 0.608 & 0.570 & 0.465 & 0.418 & 0.533 & 0.531 \\
\multirow{-4}{*}{\textbf{GraphCoT}} & DeepSeek-Chat & 0.441 & 0.443 & 0.563 & 0.466 & 0.628 & 0.546 & 0.470 & 0.418 & {\ul 0.555} & {\ul 0.559} \\ \hline
{\color[HTML]{1F1F1F} } & GPT-4o & 0.378 & 0.107 & 0.463 & 0.119 & 0.516 & 0.138 & 0.440 & 0.092 & 0.344 & 0.087 \\
{\color[HTML]{1F1F1F} } & GLM-4.6 & 0.467 & 0.101 & 0.504 & 0.090 & 0.593 & 0.129 & 0.445 & 0.090 & 0.356 & 0.079 \\

{\color[HTML]{1F1F1F} } & Qwen-Max & 0.341 & 0.104 & 0.471 & 0.100 & 0.499 & 0.153 & 0.430 & 0.098 & 0.317 & 0.074 \\
\multirow{-4}{*}{{\color[HTML]{1F1F1F} \textbf{PolyG}}} & DeepSeek-Chat & 0.493 & 0.104 & 0.533 & 0.110 & 0.585 & 0.146 & 0.445 & 0.087 & 0.339 & 0.070 \\ \hline
 & GPT-4o & 0.485 & 0.441 & 0.592 & 0.438 & 0.644 & 0.605 & 0.490 & 0.381 & 0.433 & 0.219 \\
 & GLM-4.6 & 0.407 & 0.421 & 0.579 & 0.457 & 0.595 & 0.589 & 0.490 & 0.459 & 0.339 & 0.246 \\

 & Qwen-Max & {\ul 0.530} & {\ul 0.484} & {\ul 0.625} & 0.524 & 0.637 & 0.639 & 0.459 & 0.447 & 0.478 & 0.262 \\
\multirow{-4}{*}{\textbf{GraphCounselor}} & DeepSeek-Chat & 0.470 & 0.452 & 0.621 & {\ul 0.578} & {\ul 0.657} & {\ul 0.659} & 0.510 & {\ul 0.489} & 0.483 & 0.286 \\ \hline
  & 4B w/o train & 0.211 & 0.217 & 0.179 & 0.148 & 0.210 & 0.220 & 0.250 & 0.255 & 0.189 & 0.226 \\ 
 & 4B w/ train & { \textbf{0.819}} & { \textbf{0.855}} & { \textbf{0.696}} & { \textbf{0.646}} & { \textbf{0.710}} & { \textbf{0.739}} & { 0.575} & { 0.562} & { \textbf{0.567}} & { \textbf{0.597}} \\ \cline{2-12} 
 & 8B w/o train & {\color[HTML]{1F1F1F} 0.403} & {\color[HTML]{1F1F1F} 0.415} & 0.396 & 0.386 & 0.422 & 0.401 & 0.365 & 0.358 & 0.361 & 0.364 \\
\multirow{-4}{*}{\textbf{GraphScout}} & 8B w/ train & {\color[HTML]{1F1F1F}  0.770} & {\color[HTML]{1F1F1F}  0.783} & {\color[HTML]{1F1F1F}  0.663} & {\color[HTML]{1F1F1F}  0.628} & {0.685} & { 0.702} & {\color[HTML]{1F1F1F} { \textbf{0.585}}} & {\color[HTML]{1F1F1F} { \textbf{0.569}}} & 0.555 & {0.573} \\ 
\hline
\multicolumn{2}{c}{\textbf{Relative Improvement}} &
\textbf{54.6\%} & \textbf{76.5\%} &
\textbf{11.4\%} & \textbf{11.8\%} &
\textbf{8.2\%} & \textbf{12.0\%} &
\textbf{7.3\%} & \textbf{16.3\%} &
\textbf{2.2\%} & \textbf{6.8\%} \\
\bottomrule
\end{tabular}
\end{table*}

\subsubsection{\textbf{Reward design}}
\label{sec:navi_reward}
Graph Solver is trained with a trajectory-level reward that combines (i) answer correctness and (ii) evidence clues alignment, with strict format gating. 
\zty{First, we extract the output answer $\hat{y}$ from the final trajectory $\tau$, and then retrieve the corresponding node set  of evidence clues $c \in \mathcal{C}_\mathrm{train}$ (i.e., \texttt{clue\_nodes}) collected by Graph Quizzer for reward computation. 
}

\textbf{Answer reward.}
We first extract the predicted answer $\hat{y}$ from a required wrapper \texttt{\textbackslash answer\{...\}}. If the wrapper is missing, the reward is set to zero:
\begin{equation}
\label{eq.gating} 
\hat{y} = \mathrm{ExtractAnswer}(\tau), \qquad
\hat{y}=\emptyset \ \Rightarrow\ r(\tau)=0.
\end{equation}
For valid outputs from the trajectory $\tau$, we compute an answer reward using an F1-based metric:
\begin{equation}
\label{eq.main_score} 
r_{\text{ans}}(\tau) = \mathrm{F1}(\hat{y}, y)\in[0,1].
\end{equation}
The answer reward directly measures answer-level correctness and serves as the primary training signal, ensuring that the optimization objective remains aligned with the final task performance, where the F1 score measures token-level overlap.

\textbf{Evidence clues-based reward.}
To guide exploration behavior, we introduce an evidence clues-based process reward. Let $\mathbb{I}(\tau, e_i)$ indicate whether the model’s trajectory $\tau$ references or interacts with evidence node $e_i$. The evidence reward is then defined as:
\begin{equation}
\label{eq.evid} 
r_{\text{clue}}(\tau) = \frac{1}{|c|} \sum_{i=1}^{|c|} \mathbb{I}(\tau, e_i).
\end{equation}
\zty{Relying solely on answer-based rewards yields sparse supervision signals for multi-turn graph reasoning. 
An auxiliary evidence-based reward provides denser and graph-grounded intermediate supervision that guides the model’s intermediate steps.
}

The final reward follows a case-based rule that applies the evidence reward \emph{only} when outcome quality is low and caps the combined score in this regime:
\begin{equation}
r(\tau)=
\begin{cases}
0, & \hat{y}=\emptyset ,\\[4pt]
r_{\text{ans}}(\tau), &  \ r_{\text{ans}}(\tau)\ge \delta,\\[6pt]
\min\!\big(r_{\text{ans}}(\tau)+r_{\text{clue}}(\tau),\ \delta\big), &
 \ r_{\text{ans}}(\tau)< \delta,
\end{cases}
\end{equation}
where $\delta$ is a threshold used to identify low-accuracy predictions and also serves as the upper bound for the bonus-augmented regime. This design ensures that (i) malformed outputs receive no reward, (ii) answer correctness remains the primary training signal, and (iii) evidence alignment provides auxiliary guidance without allowing low-accuracy trajectories to obtain  high rewards.

\subsubsection{\textbf{Optimization}}
Graph Solver is trained using multi-turn reinforcement learning, where the model performs multiple reasoning and tool-use steps before producing a final answer. We employ Group Relative Policy Optimization (GRPO)~\cite{guo2025deepseekr1} to optimize the policy. For each question $q$, we sample a group of tool-mediated interaction trajectories $G=\{\tau_1,\dots,\tau_{|G|}\}$ from the current policy $\pi_\theta$. Each trajectory $\tau_i$ receives a scalar reward $r_i$ computed as described in Section~\ref{sec:navi_reward}, while intermediate tool execution results do not directly contribute to the loss. For each trajectory $\tau_i$, we compute a normalized relative advantage
$A_i = (r_i - \mathrm{mean}\{r_k\}_{k=1}^{|G|}) / \mathrm{std}\{r_k\}_{k=1}^{|G|}$.

The policy parameters $\theta$ are then updated by maximizing a clipped surrogate objective with a KL penalty:
\begin{equation}
\label{eq.grpo}
L(\theta) =
\frac{1}{|G|}\sum_{i=1}^{|G|}
\Big[
\min\big(\rho_i A_i,\; \mathrm{clip}(\rho_i,1-\varepsilon,1+\varepsilon)A_i\big)
- \beta\, D_{\mathrm{KL}}(\pi_\theta \,\|\, \pi_{\text{ref}})
\Big],
\end{equation}
where $\rho_i = \pi_\theta(\tau_i \mid q) / \pi_{\theta_{\text{old}}}(\tau_i \mid q)$ is the importance ratio computed from model-generated tokens, $\varepsilon$ controls the clipping range, and $\beta$ weights the KL regularization term that constrains deviation from the reference policy $\pi_{\text{ref}}$. More details are provided in Appendix~\ref{app:grpo}.

\section{Experiment}
\subsection{Experimental Setup}
\textbf{Dataset.}
We conduct experiments on GRBENCH dataset~\cite{jin2024graphcot} following baseline approaches~\cite{jin2024graphcot,gao2025graphCounselor}, which contains graphs from five domains and a total of 1,740 questions categorized into three difficulty levels: easy, medium, and hard. All questions are in English. More details can be found in the Appendix~\ref{app:dataset_details}.\\
\textbf{Baselines.}
We compare our GraphScout approach with several baselines: BaseLLM, TextRAG~\cite{gao2023rag3}, GraphRAG~\cite{ye2024language}, Cypher~\cite{liu2025polyg}, GraphCoT~\cite{jin2024graphcot}, PolyG~\cite{liu2025polyg}, and GraphCounselor~\cite{gao2025graphCounselor}.
To ensure a comprehensive and fair evaluation of these baselines, we instantiate each method with multiple state-of-the-art LLMs, including  GPT-4o~\cite{achiam2023gpt4}, GLM-4.6~\cite{glm2024chatglm}, Qwen-Max~\cite{yang2025qwen3}, and DeepSeek-Chat~\cite{liu2024deepseek}. Detailed settings are provided in the Appendix~\ref{sec:baseline_details}.\\
\textbf{Evaluation Metrics.}
We evaluate model performance using both model-based and automatic metrics. 
Specifically, we employ QwenScore, an LLM-based metric using Qwen-Max as a judge to assess answer correctness by comparing generated outputs with ground-truth answers, and report the proportion of responses judged as correct.
Besides, we report the F1 score, which measures token-level overlap between predicted and ground-truth answers.\\
\textbf{Models and Training Settings.}
We conduct experiments using two model variants: Qwen3-4B-Instruct-2507 and Qwen3-8B. For the Graph Quizzer stage, we use DeepSeek-Chat to explore the knowledge graph and synthesize sufficient training data. To prevent test-set leakage, we ensure that all clue nodes in the training set are disjoint from any nodes appearing in the test set questions. During training, we adopt the \texttt{verl}~\cite{sheng2024verl} framework and employ its implementation of Group Relative Policy Optimization (GRPO)~\cite{guo2025deepseekr1} for reinforcement learning, training the models for 400 optimization steps. Detailed training hyperparameters and configurations are provided in the Appendix~\ref{app:rlhyper}.

\begin{figure}[]
    \centering 
    \includegraphics[width=1\columnwidth]{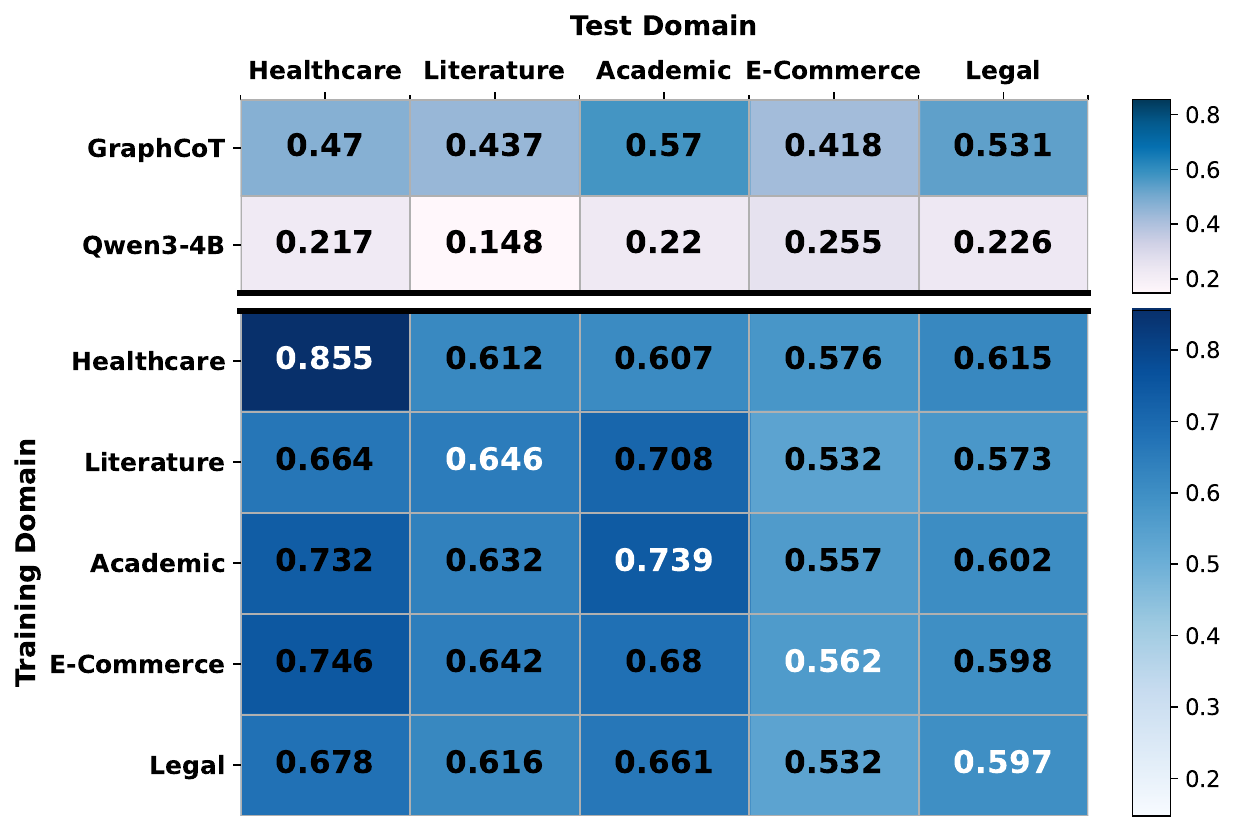}
 \caption{Cross-Domain Generalization Performance Across Training and Test Domains.}
    \label{fig:general}
\end{figure}

\begin{figure*}[h]
    \centering 
    \includegraphics[width=1\linewidth]{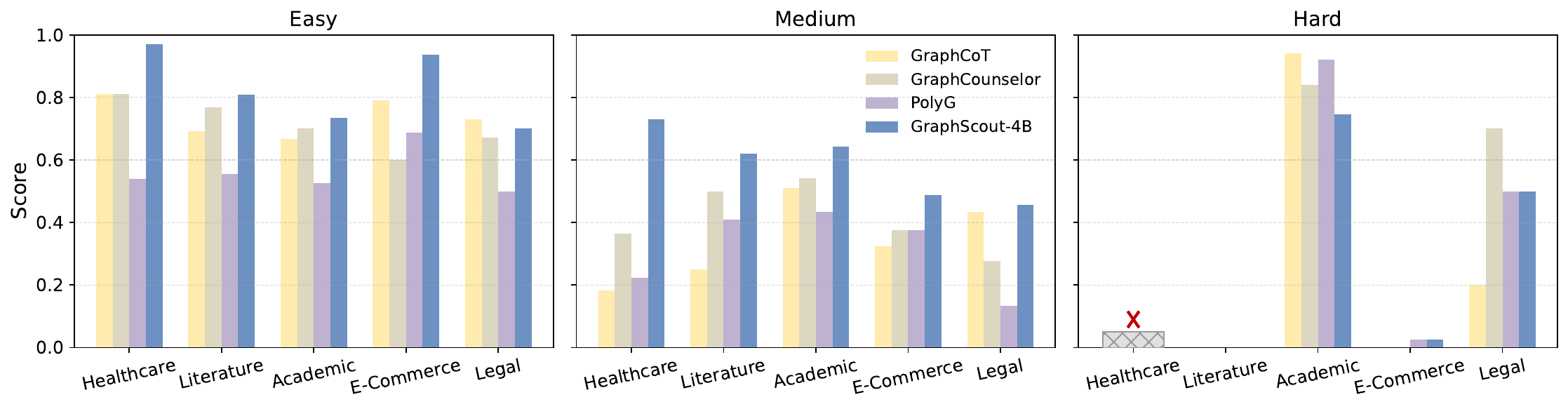}
\caption{Performance Comparison Across Question Difficulty Levels on GRBENCH using F1-score as the metric. The Healthcare domain contains no hard questions. For the Literature, all methods achieve an F1-score of 0\% on hard questions.}
    \label{fig:difficulty}
\end{figure*}

\subsection{Performance Analysis}
\subsubsection{\textbf{Overall Accuracy  Analysis}}
Table~\ref{table:main} reports performance on GRBENCH across five domains using both QwenScore and F1. Overall, graph-based approaches consistently outperform BaseLLM and TextRAG, underscoring the importance of explicitly leveraging structured graph knowledge. Methods such as Cypher and PolyG achieve relatively high QwenScore but noticeably lower F1, which is expected as they do not enforce strict answer format constraints. In such cases, QwenScore, which emphasizes semantic equivalence, better reflects answer quality.

Despite relying on substantially smaller backbone LLMs, GraphScout achieves the strongest overall performance after post-training, surpassing all baseline methods built upon leading LLMs such as DeepSeek-Chat and Qwen-Max. This advantage is particularly pronounced on the Healthcare dataset, which features a highly complex graph structure with 11 node types and 24 edge types. Under such conditions, baseline methods, even when supported by powerful LLM backbones, are constrained by limited graph exploration capabilities and thus struggle to reason effectively over intricate relational structures. In contrast, GraphScout is explicitly designed to support agentic graph reasoning by equipping models with intrinsic exploration ability and flexible graph interaction tools, making it especially well suited to reasoning over complex graphs. As a result, it demonstrates particularly strong performance in this dataset. Case studies are provided in Appendix~\ref{app:case_study}.

Regarding model scale, GraphScout-8B outperforms GraphScout-4B without training. After post-training, however, the 4B variant achieves stronger performance than the 8B model in most cases, consistent with observations from prior post-training studies~\cite{distillab2025}. We hypothesize that this behavior may be related to the hybrid reasoning configuration of Qwen3-8B.

\subsubsection{\textbf{Cross-Domain Generalization Analysis}}

We further evaluate the robustness of GraphScout under cross-domain settings to examine whether the learned exploration capability transfers beyond the training domain. Figure~\ref{fig:general} reports the F1-score performance of GraphScout-4B when trained on a single domain and evaluated on all five GRBENCH domains. For comparison, we also include the performance of Qwen-Max based GraphCoT and the base Qwen3-4B-Instruct evaluated independently on each domain.

Overall, GraphScout exhibits robust cross-domain performance across all training–testing combinations, substantially outperforming the baselines in most cases. Notably, models trained on one domain consistently achieve solid performance on unseen domains, with only mild performance degradation under domain shifts. This indicates that GraphScout learns transferable, intrinsic graph exploration behaviors rather than memorizing domain-specific structures or traversal heuristics, enabling effective adaptation across heterogeneous graphs without additional efforts.

\begin{table}[]
\caption{Ablation study of GraphScout (``rw'' denotes random walk on knowledge graphs).}
\label{tb:ablation}
\setlength{\tabcolsep}{4.5pt} 
\begin{tabular}{ccccc}
\toprule
 & \multicolumn{2}{c}{Healthcare} & \multicolumn{2}{c}{Literature} \\ \cline{2-5} 
 & QwenScore & F1 & QwenScore & F1 \\ \hline
GraphScout & {\color[HTML]{000000} \textbf{0.819}} & {\color[HTML]{000000} \textbf{0.855}} & {\color[HTML]{000000} \textbf{0.696}} & {\color[HTML]{000000} \textbf{0.646}} \\
w/o Graph Solver & {\color[HTML]{1F1F1F} 0.211} & {\color[HTML]{1F1F1F} 0.217} & 0.210 & 0.220 \\
w/o $\mathcal{A}_\mathrm{code}$ & 0.107 & 0.101 & 0.196 & 0.148 \\
w/o $r_\mathrm{clue}$ & 0.785 & 0.812 & 0.667 & 0.635 \\
rw Graph Quizzer & 0.678 & 0.705 &0.608  & 0.571 \\ \bottomrule
\end{tabular}
\end{table}

\subsubsection{\textbf{Ablation Analysis}}
Table~\ref{tb:ablation} reports an ablation study of GraphScout. Removing Graph Solver (\emph{w/o Graph Solver}) leads to substantial performance drops, confirming that the gains stem from learned exploration behaviors rather than pretrained knowledge alone. Eliminating Code Interpreter (\emph{w/o $\mathcal{A}_\mathrm{code}$}) causes the most severe degradation, highlighting the necessity of tool-mediated graph interaction.
Removing Evidence reward (\emph{w/o $r_\mathrm{clue}$}) consistently reduces performance, indicating that graph evidence provides important process-level guidance beyond final answer correctness. Finally, replacing Graph Quizzer with random-walk-based question generation~\cite{lu2025deepdive} (\emph{rw Graph Quizzer}) also degrades performance. Here, training data are generated by sampling subgraphs through random walks, converting them into text, and prompting an LLM to produce question–answer pairs without interactive exploration. The performance drop suggests that supervision derived from static subgraph sampling provides weaker guidance for learning effective exploration behaviors.

\subsubsection{\textbf{Accuracy Across Question Difficulty Levels}}
To better understand how GraphScout behaves under different reasoning requirements, we further break down the GRBENCH results by difficulty level (Figure~\ref{fig:difficulty}). In GRBENCH, \textbf{Easy} questions mainly test basic fact retrieval and single-step reasoning; \textbf{Medium} questions require multi-step traversal, more complex path discovery, and evidence aggregation; \textbf{Hard} questions emphasize more inductive reasoning and are often recommendation-like. 

\textbf{Easy.} GraphScout-4B consistently achieves strong performance across all five domains on easy questions, often outperforming the strongest graph-based baselines. This suggests that the model learns to reliably ground answers on graph information for common single-hop or shallow retrieval queries.

\textbf{Medium.} The advantage of GraphScout becomes even more evident on medium difficulty. Across domains, GraphScout-4B outperforms the baselines by a substantial margin, indicating strong capability in multi-turn exploration and multi-hop information aggregation. These results align with our motivation that learning intrinsic exploration ability (rather than relying on prompting heuristics) is particularly beneficial when the question requires iterative tool use and compositional reasoning.

\textbf{Hard.} Hard questions are challenging for all methods. Many hard instances in this dataset emphasize recommendation-style reasoning and external world knowledge, placing less emphasis on structured graph traversal. Consequently, the benefits of stronger graph exploration ability are less pronounced, and GraphScout-4B shows limited gains on this split.

\begin{figure}[h]
    \centering 
    \includegraphics[width=1\columnwidth]{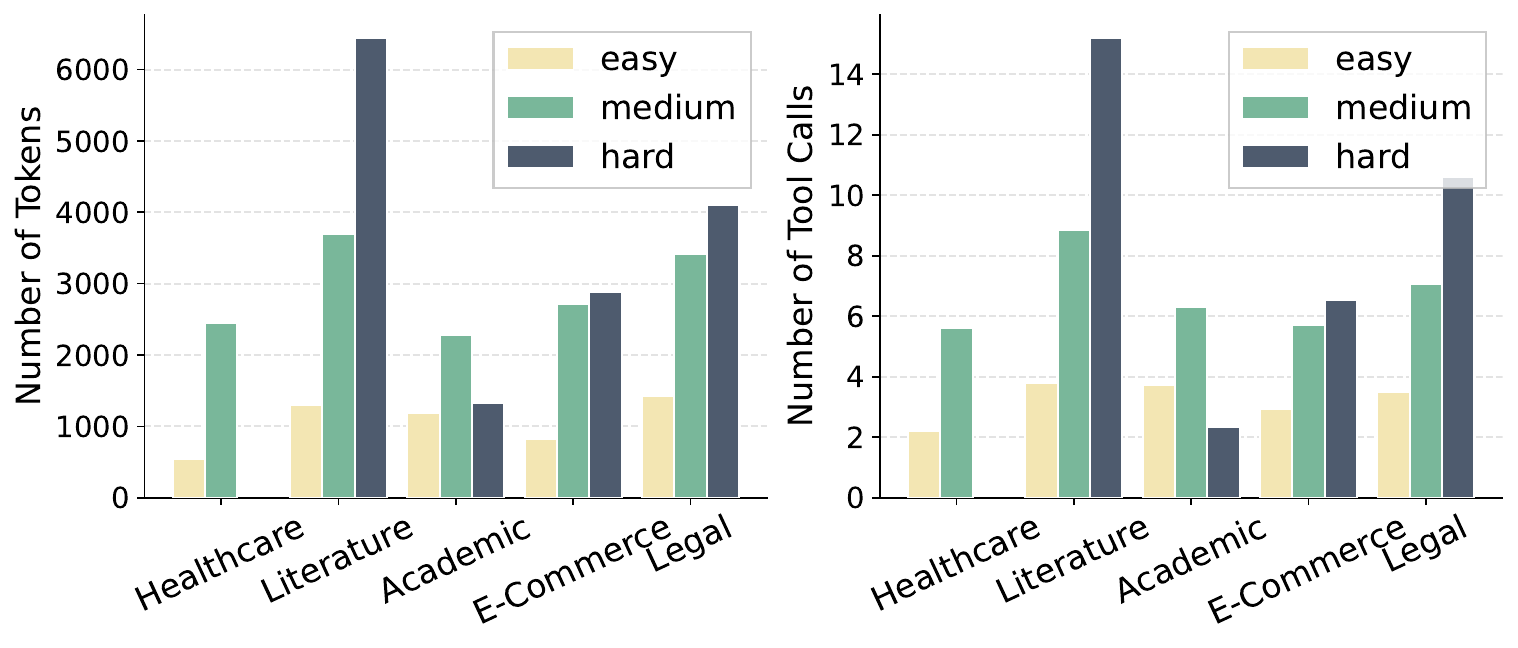}
\caption{Behavioral analysis across question difficulty levels. }
    \label{fig:token_call}
\end{figure}

\subsubsection{\textbf{Reasoning Behavior Across Question Difficulty Levels}}
Figure~\ref{fig:token_call} reports GraphScout’s behavior across question difficulty levels using average output length and tool calls per question. In most domains, both metrics increase from easy to hard, suggesting deeper reasoning and more extensive graph interaction as task complexity rises. 
An exception occurs in the Academic domain for hard questions: these instances primarily involve recommendations within a constrained publication scope, emphasizing semantic understanding rather than explicit graph querying.

\begin{figure}[h]
    \centering 
    \includegraphics[width=0.95\columnwidth]{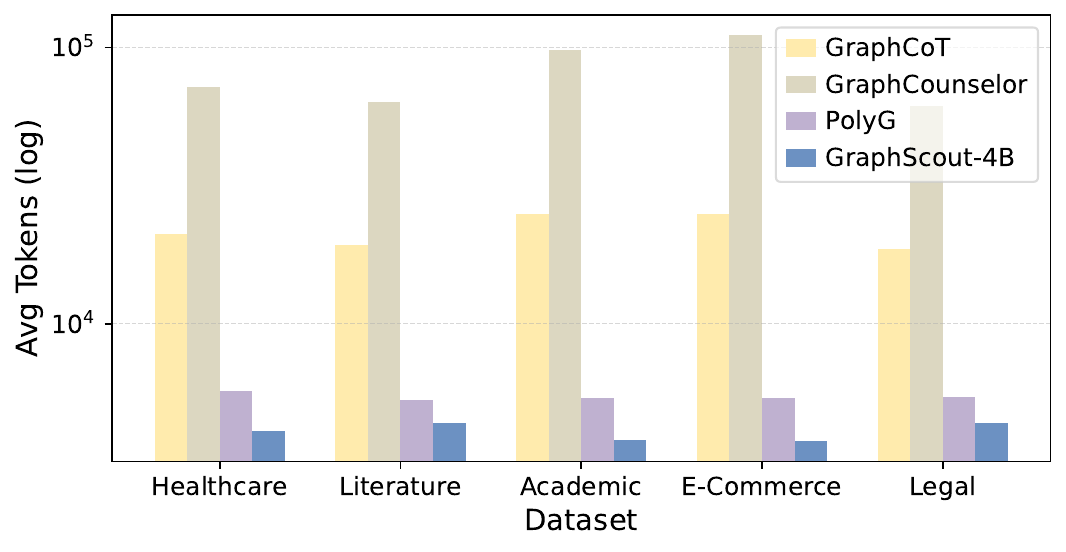}
 \caption{Average token consumption of different methods.}
    \label{fig:token}
\end{figure}

\subsubsection{\textbf{Efficiency Analysis}}
Figure~\ref{fig:token} compares the average token consumption of different methods across GRBENCH domains. Overall, GraphScout-4B consistently uses substantially fewer tokens than other active traversal-based baselines. Importantly, this reduction does not come at the cost of performance; as shown in previous results, GraphScout achieves  superior accuracy while maintaining significantly lower computational overhead.


\section{Discussion}
To position GraphScout, we contrast two representative GraphRAG settings: \textbf{Document-Centric}, which constructs graphs on-the-fly from unstructured text corpora (representative examples include HippoRAG~\cite{gutierrez2024hipporag}, HyperGraphRAG~\cite{luo2025hypergraphrag}, and Graph-R1~\cite{luo2025graphr1}), and \textbf{Native-KG Reasoning}, which reasons over pre-existing curated knowledge graphs (representative examples include GraphCoT~\cite{jin2024graphcot}, GraphCounselor~\cite{gao2025graphCounselor}, and our proposed GraphScout). These settings differ in several respects. First, they assume different graph sources: document-centric methods induce graphs from text, whereas Native KG methods operate on persistent curated graphs. Second, their main bottlenecks are different. Document-centric methods are often bottlenecked by graph construction quality, information loss during text-to-graph conversion, and effective multi-hop retrieval over noisy induced structures. In contrast, Native KG reasoning must jointly satisfy topological constraints and semantic grounding constraints. Third, their technical goals are also different: document-centric methods mainly improve retrieval and evidence connectivity across textual silos, whereas native KG methods target accurate multi-hop reasoning over explicit relational structure. Since these settings involve different inputs, failure modes, and evaluation priorities, we focus our empirical comparisons on baselines under the Native KG reasoning setting, which is the setting directly addressed by GraphScout.

\section{Conclusion}

\zty{This paper presents \NAME, a framework that enhances LLMs' agentic graph reasoning with intrinsic exploration ability.
\NAME consists of three core components: Agentic Graph Exploration Tools, which equip LLMs with more flexible and programmable graph interaction interface; Graph Quizzer, which explores diverse and high-quality problem sets on graphs while recording traversal clues; and Graph Solver, which post-trains LLMs to acquire intrinsic graph exploration ability. 
Extensive experiments against \emph{state-of-the-art} GraphRAG methods demonstrate that small-parameter LLMs equipped with \NAME significantly outperforms substantially larger LLMs by an average of 16.7\%.
Moreover, \NAME exhibits promising cross-domain generalization capability, indicating that LLMs can develop generalized graph reasoning ability even with training on a single graph.
Future work will investigate LLMs bootstrapping via iterative self-questioning and self-answering in an iterative enhancement loop, while addressing challenges such as training instability and self-bias amplification.

}

\bibliographystyle{ACM-Reference-Format}
\bibliography{references}

\appendix

\section{Implementation Details}

\subsection{RL Hyperparameters}
\label{app:rlhyper}

The experimental settings~\cite{sheng2024verl} for training Base model using our GraphScout framework are shown in Table~\ref{tab:grpo_settings}. 

\begin{table}[h]
\centering
\caption{GRPO Training Settings for GraphScout.}
\label{tab:grpo_settings}
\setlength{\tabcolsep}{10pt}
\begin{tabular}{ll}
\toprule
\textbf{Method} & \textbf{Hyperparameters} \\
\midrule

GraphScout
&
\begin{tabular}[t]{@{}l@{}}
RL Algorithm = GRPO \\
n\_resp\_per\_prompt = 8 \\
Temperature = 1.0 \\
Threshold $\delta$ = 0.4 \\
Optimization Steps = 400 \\
Training samples per dataset = 2000 \\
\end{tabular}
\\

\midrule
Backbone
&
\begin{tabular}[t]{@{}l@{}}
Qwen3-4B-Instruct-2507, \\
Qwen3-8B \\
\end{tabular}
\\

\midrule
GRPO Trainer
&
\begin{tabular}[t]{@{}l@{}}
Actor learning rate = $1\times10^{-6}$ \\
clip\_ratio = [0.20, 0.28] \\
Initial KL Coefficient = $1\times10^{-4}$ \\
\end{tabular}
\\

\midrule
Batch Sizes
&
\begin{tabular}[t]{@{}l@{}}
train\_batch\_size = 16 \\
rollout\_batch\_size = 16 \\
micro\_train\_batch\_size = 8 \\
micro\_rollout\_batch\_size = 16 \\
\end{tabular}
\\

\midrule
Lengths
&
\begin{tabular}[t]{@{}l@{}}
Prompt Max Length = 4096 \\
Generate Max Length = 8192 \\
\end{tabular}
\\

\midrule
Optimizations
&
\begin{tabular}[t]{@{}l@{}}
bf16, gradient\_checkpointing, \\
 remove\_padding, packing\_samples, \\
 dynamic\_batch\_size, flash\_attn  \\
\end{tabular}
\\

\bottomrule
\end{tabular}
\end{table}

\subsection{Hardware and Software Configuration}
All experiments were conducted on a Linux server running Ubuntu 20.04.5 LTS.
The server is equipped with 8 NVIDIA A800-SXM4 GPUs (80 GB memory per GPU) and powered by dual Intel Xeon Platinum 8358 CPUs  with 1 TB system memory.
The system runs CUDA 12.4, and all post-training experiments were conducted using the verl framework~\cite{sheng2024verl}.

\subsection{Details of Dataset}
\label{app:dataset_details}
\textbf{Dataset Overview.}
GRBENCH is a graph reasoning benchmark designed to evaluate how LLMs interact with large-scale, text-attributed graphs.
It consists of ten real-world graphs spanning five domains: academic, e-commerce, literature, healthcare, and legal.
In total, GRBENCH contains 1,740 manually designed question-answer pairs, all in English, covering three difficulty levels: easy, medium, and hard.
Each question can be answered by explicitly reasoning over the corresponding domain graph rather than relying solely on parametric knowledge. All dataset statistics and construction details follow the original GRBENCH benchmark~\cite{jin2024graphcot}.

\begin{table}[h]
\centering
\caption{Dataset Statistics of GRBENCH.}
\label{tab:grbench_stats}
\small
\setlength{\tabcolsep}{3pt}
\begin{tabular}{llcccc}
\toprule
\textbf{Domain} & \textbf{Topic} & \textbf{\# Nodes} & \textbf{\# Edges} & \textbf{\# Temp} & \textbf{\# Q} \\
\midrule
\multirow{6}{*}{Academic}
 & CS                & $\sim$8M  & $\sim$52M  & 15 & 150 \\
 & Biology           & $\sim$4M  & $\sim$39M  & 14 & 140 \\
 & Chemistry         & $\sim$4M  & $\sim$30M  & 14 & 140 \\
 & Material Science  & $\sim$3M  & $\sim$22M  & 14 & 140 \\
 & Medicine          & $\sim$6M  & $\sim$30M  & 14 & 140 \\
 & Physics           & $\sim$2M  & $\sim$33M  & 14 & 140 \\
\midrule
E-commerce & Amazon     & $\sim$9M  & $\sim$313M & 20 & 200 \\
Literature & Goodreads  & $\sim$3M  & $\sim$22M  & 24 & 240 \\
Healthcare & Disease    & $\sim$47K & $\sim$4M   & 27 & 270 \\
Legal      & Freelaw    & $\sim$84M & $\sim$114M & 18 & 180 \\
\midrule
\textbf{SUM} & -- & -- & -- & \textbf{174} & \textbf{1740} \\
\bottomrule
\end{tabular}
\end{table}

\textbf{Academic Domain.}
The academic domain consists of six large bibliographic graphs covering Computer Science, Biology, Chemistry, Material Science, Medicine, and Physics.
These graphs are constructed from DBLP~\cite{tang2008dblp} and Microsoft Academic Graph (MAG)~\cite{wang2020mag}.
Nodes include papers, authors, and venues, while edges represent citation relations, authorship relations, and publication venues.
The academic graphs support multi-hop reasoning over citation chains, co-authorship structures, and publication histories.

\textbf{E-commerce Domain.}
The e-commerce domain is built from the Amazon product datasets~\cite{he2016amazon}.
Nodes correspond to items and brands, and edges capture user behavior signals such as \emph{also-viewed}, \emph{also-bought}, \emph{buy-after-viewing}, and \emph{bought-together}.
This domain emphasizes large-scale, densely connected graphs and supports aggregation-based reasoning tasks.

\textbf{Literature Domain.}
The literature domain is based on the Goodreads dataset~\cite{wan2018goodreads}.
It contains nodes such as books, authors, publishers, and series, with edges describing authorship, publication, series membership, and book similarity.
Questions in this domain often require reasoning over content attributes  as well as relational structures among books and authors.

\textbf{Healthcare Domain.}
The healthcare domain adopts the Hetionet biomedical knowledge graph~\cite{himmelstein2017biomedical}.
It is a highly heterogeneous graph consisting of eleven node types, including diseases, compounds, genes, symptoms, and side effects.
Edges describe rich biomedical relations.
This domain supports complex biomedical reasoning that involves aggregating evidence across multiple relation types.

\textbf{Legal Domain.}
The legal domain is constructed from CourtListener data.
Nodes include opinions, opinion clusters, dockets, and courts, while edges encode citation relations and hierarchical judicial structures.
Legal questions often require tracing citation paths or aggregating information across multiple legal documents.

\subsection{Details of Baseline}
\label{sec:baseline_details}

In this section, we provide detailed descriptions of all baseline methods evaluated in our experiments, including their underlying assumptions, graph interaction mechanisms, and reasoning paradigms. All baselines are implemented following their original papers and, unless otherwise specified, are evaluated under the same  evaluation protocols as GraphScout to ensure fair comparison.

\textbf{BaseLLM.}
BaseLLM serves as a non-graph baseline to assess the extent to which LLMs can answer domain-specific questions relying solely on their internal parametric knowledge. For this baseline, the model is prompted directly with the question, without access to any external graph or retrieved context. 

\textbf{TextRAG.}
TextRAG~\cite{gao2023rag3} treats the knowledge graph as a textual corpus by linearizing node attributes and edge information into plain text. A dense retriever is employed to fetch the most relevant text chunks based on semantic similarity with the input question, and the retrieved text is appended to the prompt for answer generation.

\textbf{GraphRAG.}
GraphRAG~\cite{ye2024language} extends TextRAG by incorporating structural information during retrieval. Instead of retrieving isolated text chunks, GraphRAG retrieves subgraphs centered around core entities identified from the question, typically using 2-hop neighborhood expansion. The retrieved subgraphs are then linearized into text sequences and provided as context to the LLM. 

\textbf{Cypher.}
The Cypher baseline directly queries the underlying knowledge graph using Cypher. For each question, the LLM is prompted to generate a Cypher query, which is executed against the graph to retrieve relevant nodes or attributes. The query results are then returned to the LLM for answer synthesis. To adapt this method to the tasks considered in this paper, we additionally incorporate a node retrieval module to identify candidate nodes prior to Cypher query execution.

\textbf{GraphCoT.}
GraphCoT~\cite{jin2024graphcot} is a reasoning-based GraphRAG method that enables multi-step interaction between the LLM and the graph through a predefined set of graph traversal tools. The model incrementally explores the graph, retrieves intermediate evidence, and updates its reasoning context in a chain-of-thought manner. By decomposing complex queries into multiple reasoning steps, GraphCoT mitigates information loss caused by long linearized contexts and serves as a strong iterative reasoning baseline.

\textbf{PolyG.}
PolyG~\cite{liu2025polyg} introduces an adaptive graph traversal strategy that dynamically adjusts exploration paths based on the evolving reasoning state. Unlike fixed-hop retrieval methods, PolyG allows the model to select different graph operators conditioned on the query and intermediate observations. This flexibility improves coverage for diverse question types, particularly those requiring non-uniform traversal patterns. To adapt this method to the tasks considered in this paper, we additionally incorporate a node retrieval module to identify candidate nodes prior to graph traversal.

\textbf{GraphCounselor.}
GraphCounselor~\cite{gao2025graphCounselor} employs a multi-agent collaborative framework for graph reasoning. It decomposes the reasoning process into planning, execution, and reflection stages, allowing the model to dynamically adjust reasoning depth and correct earlier mistakes through self-reflection. 

\textbf{LLM Backbones.}
To ensure robustness and generalizability, all baselines are instantiated with multiple state-of-the-art foundation models, including GPT-4o~\cite{achiam2023gpt4}, GLM-4.6~\cite{glm2024chatglm},  Qwen-Max~\cite{yang2025qwen3}, and DeepSeek-Chat~\cite{liu2024deepseek}. Unless otherwise noted, decoding parameters (e.g., temperature and top-$p$) follow the default or recommended settings from the corresponding baseline implementations. In addition, for all methods that require vector retrieval, we use \texttt{Qwen3-text-embedding-v4} to obtain text embeddings.

Overall, these baselines cover a broad spectrum of graph reasoning paradigms, ranging from text-only retrieval and symbolic querying to iterative and adaptive graph exploration.

\begin{figure*}[h]
    \centering 
    \includegraphics[width=1\linewidth]{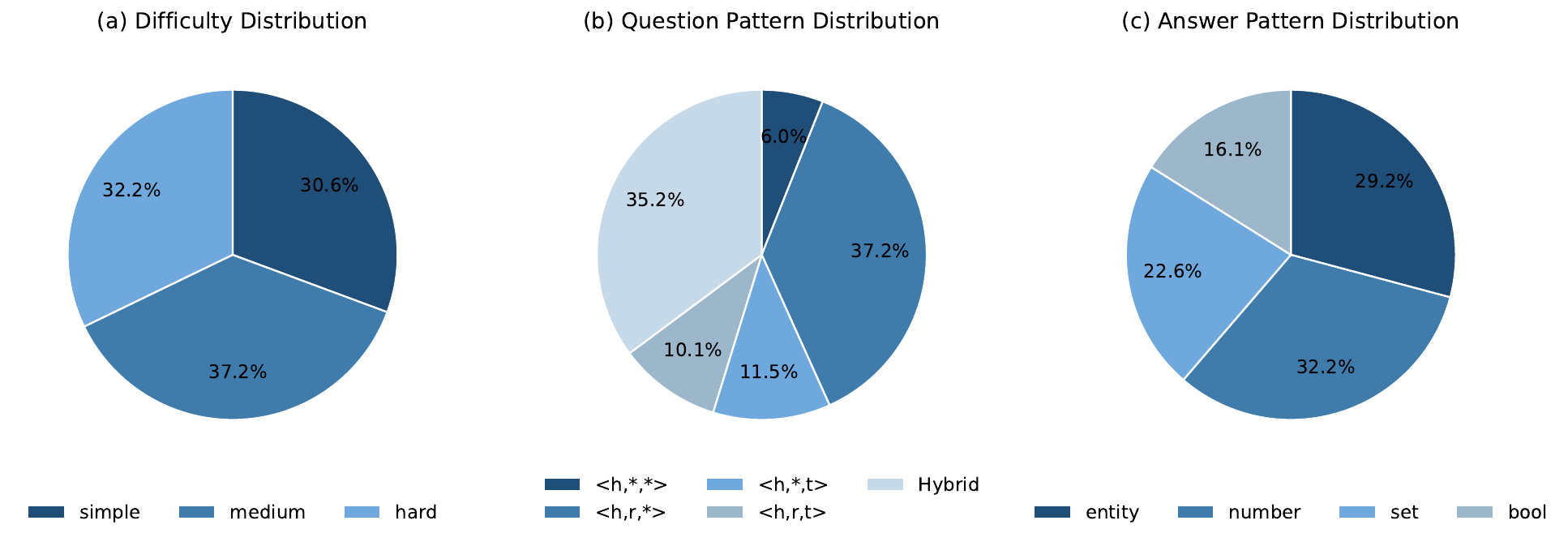}
\caption{Distribution of the generated dataset in terms of question difficulty levels, question structural patterns, and answer types. All category labels are automatically judged by an LLM. The three pie charts summarize the proportions of different categories across the entire dataset.}
    \label{fig:Diversity}
\end{figure*}

\subsection{\textbf{Group Relative Policy Optimization}}
\label{app:grpo}

We adopt Group Relative Policy Optimization (GRPO)~\cite{guo2025deepseekr1} as the reinforcement learning algorithm for training the graph reasoning policy. GRPO is a variant of Proximal Policy Optimization (PPO) that eliminates the need for an explicit value function by estimating advantages using relative rewards within a sampled group, thereby simplifying training and reducing computational overhead.

For each input question $q$, a frozen reference policy $\pi_{\text{ref}}$ and the current policy $\pi_{\theta}$ are used to sample a group of $|G|$ trajectories:
\begin{equation}
G = \{\tau_1, \tau_2, \dots, \tau_{|G|}\}, \quad \tau_i \sim \pi_{\theta_{\text{old}}}(\cdot \mid q),
\end{equation}
where each trajectory $\tau_i$ corresponds to a complete multi-step interaction sequence, including intermediate reasoning steps and tool invocations, and terminates with a final answer. A scalar reward $r_i$ is assigned to each trajectory based solely on the correctness and quality of the final outcome, while intermediate steps are not explicitly supervised.

Unlike PPO, which relies on a learned value function to estimate advantages, GRPO computes a group-relative advantage by normalizing rewards within the sampled group:
\begin{equation}
A_i = \frac{r_i - \mu_G}{\sigma_G}, \quad
\mu_G = \frac{1}{|G|}\sum_{k=1}^{|G|} r_k,\quad
\sigma_G = \sqrt{\frac{1}{|G|}\sum_{k=1}^{|G|}(r_k - \mu_G)^2}.
\end{equation}
This formulation encourages the policy to increase the likelihood of trajectories that outperform other candidates sampled for the same question, without requiring a separate critic network.

The policy is updated using a clipped surrogate objective analogous to PPO. Specifically, let
\begin{equation}
\rho_i(\theta) = \frac{\pi_{\theta}(\tau_i \mid q)}{\pi_{\theta_{\text{old}}}(\tau_i \mid q)}
\end{equation}
denote the importance sampling ratio computed from the token-level probabilities of the generated trajectory. The GRPO objective is defined as:
\begin{equation}
\label{eq:grpo_appendix}
\begin{aligned}
L_{\text{GRPO}}(\theta)
&=
\frac{1}{|G|}\sum_{i=1}^{|G|}
\min\!\Big(
\rho_i(\theta) A_i,\;
\mathrm{clip}\big(\rho_i(\theta), 1-\varepsilon, 1+\varepsilon\big) A_i
\Big) \\
&\quad
- \beta\, D_{\mathrm{KL}}\!\left(\pi_{\theta} \,\|\, \pi_{\text{ref}}\right).
\end{aligned}
\end{equation}
where $\varepsilon$ is the clipping hyperparameter and $\beta$ controls the strength of the KL regularization term. The KL penalty constrains the updated policy to remain close to the reference policy, stabilizing training and preventing excessive divergence during long-horizon reasoning.

\section{Additional Analysis and Experiments}

\subsection{Diversity Analysis of Graph Quizzer}

To analyze the diversity of the data generated by Graph Quizzer, we perform an additional automatic annotation analysis on the generated questions. Specifically, during this analysis, we use DeepSeek-Chat as the judging LLM, while deliberately withholding the per-question task configurations assigned during generation, including the specific combinations of difficulty levels, question structural patterns, and answer types specified for each individual question. Instead, we only provide the definitions of these categories together with each generated question, allowing the model to independently determine the task category combination that each question actually corresponds to.

This de-conditioned annotation strategy is designed to analyze the generated data from the perspective of the questions’ intrinsic semantics and structures. On the one hand, during generation, Graph Quizzer may not always strictly adhere to the predefined task instructions; on the other hand, this post hoc annotation process enables us to explicitly verify the actual task attributes reflected by each question at the semantic level.

As shown in the figure~\ref{fig:Diversity}, the distributions of question difficulty and answer type are relatively balanced, indicating good coverage of the generated data along these dimensions. In contrast, the distribution of question structural patterns is more uneven, with hybrid questions accounting for a larger proportion. We hypothesize that this phenomenon primarily arises because, when generating questions of medium and higher difficulty, simple and single-structure question patterns are often insufficient to meet the required difficulty level, thereby encouraging the model to generate hybrid questions that incorporate multiple structural elements.

\begin{figure}[t]
    \centering 
    \includegraphics[width=0.7\linewidth]{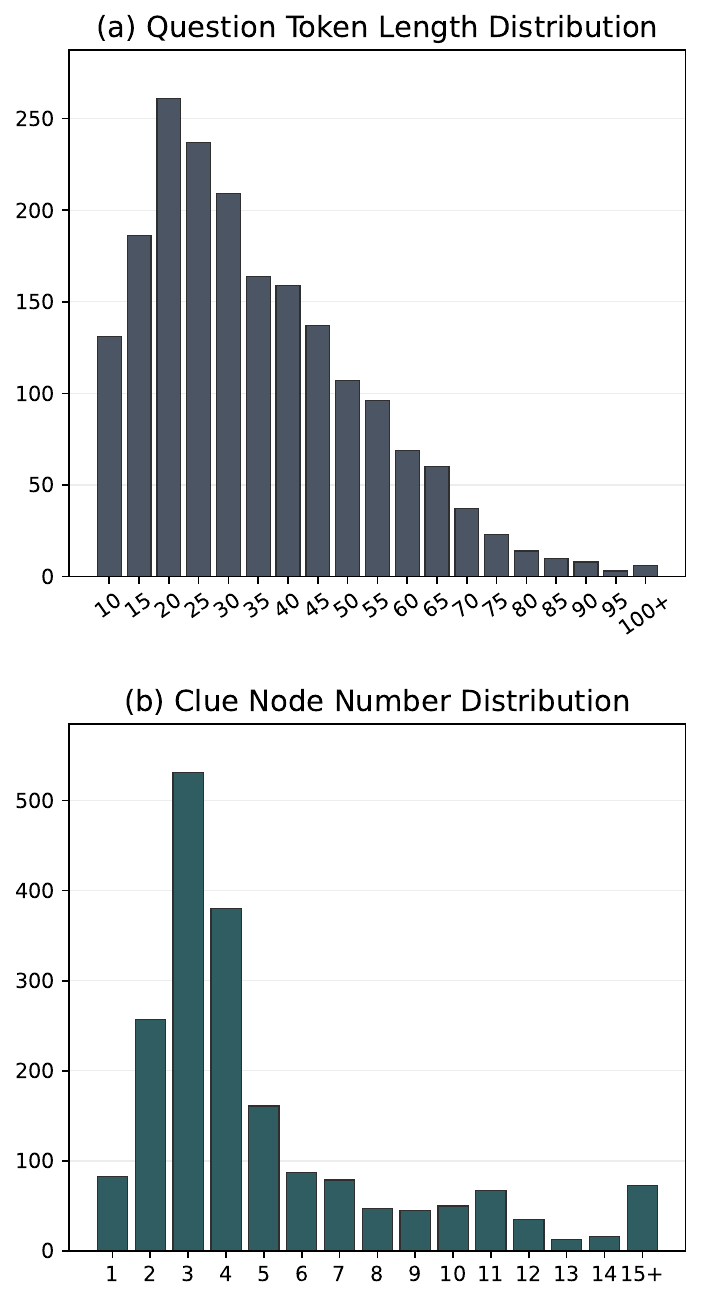}
\caption{Distribution of (a) question token lengths and (b) the number of clue nodes per question in Healthcare dataset.}
    \label{fig:tokens_clues}
\end{figure}

To further examine diversity beyond categorical task attributes, we analyze the generated data from a structural perspective in Figure~\ref{fig:tokens_clues}. Specifically, Figure~\ref{fig:tokens_clues} presents the distributions of question token lengths and the number of clue nodes per question. The token length distribution exhibits substantial variation with a clear long-tail pattern, while the number of clue nodes spans from simple, single-clue questions to instances involving multiple reasoning cues. Together, these results indicate that Graph Quizzer generates questions with diverse linguistic realizations and varying levels of reasoning complexity.

\begin{table}[h]
\caption{Proportion of failed tool calls.}
\label{tb:failed_call}
\small
\setlength{\tabcolsep}{3pt} 
\begin{tabular}{cccccc}
\toprule
 & Healthcare & Literature & Academic & E-Commerce & Legal \\ \hline
w/o train & 63.73\% & 67.14\% & 61.48\% & 61.15\% & 72.20\% \\
w/ train & \textbf{3.14\%} & \textbf{5.80\%} & \textbf{5.45\%} & \textbf{6.61\%} & \textbf{1.27\%} \\ \bottomrule
\end{tabular}
\end{table}

\subsubsection{\textbf{Tool Invocation Reliability}}
Table~\ref{tb:failed_call} reports the proportion of failed tool calls before and after training across different domains. Without training, the model exhibits a high failure rate, exceeding 60\% in most domains, indicating unreliable tool usage under open graph interfaces. After training with GraphScout, the failure rate drops sharply across all domains. This substantial reduction suggests that training on task-oriented exploration experience significantly improves the model’s ability to invoke tools correctly and robustly, highlighting the effectiveness of GraphScout in stabilizing tool-mediated graph interaction.

\subsection{Case Study}
\label{app:case_study}

\noindent\textbf{Question: } 
What cellular component is involved with most of the genes that are downregulated in a disease causing Aphasia, Primary Progressive?

\noindent\textbf{Ground Truth: } 
\emph{neuron projection.}

\subsubsection{\textbf{GraphCoT Case (Failure).}}
The model terminates with:
\emph{``Unfortunately, there is no information available in the graph about genes that are downregulated or associated with Primary Progressive Aphasia. Therefore, we cannot determine the cellular components involved with these genes for this disease.''} 

\textbf{Turn 1:}\\
\begin{safebox}
Thought 1. The question is related to a disease node (Aphasia, Primary Progressive) and we need to find the node in the graph.  
Action 1. RetrieveNode[Aphasia, Primary Progressive]  
Observation 1. The ID of this retrieval target node is D018888.
\end{safebox}

\textbf{Turn 2:}\\
\begin{safebox}
Thought 2. The next step is to find the genes that are downregulated by this disease.  
Action 2. NeighbourCheck[D018888, Disease-downregulates-Gene]  
Observation 2. The node or neighbor type does not exist in the graph. Please modify it.
\end{safebox}

\textbf{Turn 3:}\\
\begin{safebox}
Thought 3. It seems there was an issue with the neighbor type specified.  
Action 3. RetrieveNode[Primary Progressive Aphasia]  
Observation 3. The ID of this retrieval target node is D018888.
\end{safebox}

\textbf{Turn 4:}\\
\begin{safebox}
Thought 4. The node ID remains the same. Try again to find downregulated genes.  
Action 4. NeighbourCheck[D018888, Disease-downregulates-Gene]  
Observation 4. The node or neighbor type does not exist in the graph.
\end{safebox}

\textbf{Turn 5:}\\
\begin{safebox}
Thought 5. Explore alternative relations using Disease-associates-Gene.  
Action 5. NeighbourCheck[D018888, Disease-associates-Gene]  
Observation 5. The node or neighbor type does not exist in the graph.
\end{safebox}

\vspace{0.5em}

\textbf{Turn 6:}\\
\begin{safebox}
Thought 6. No associated or downregulated genes are available for this disease.  
Action 6. Finish[Unfortunately, there is no information available in the graph about genes that are downregulated or associated with Primary Progressive Aphasia. Therefore, we cannot determine the cellular components involved with these genes for this disease.]  
Observation 6. Answer is INCORRECT.
\end{safebox}

\subsubsection{\textbf{GraphScout Case (Success).}}
The model outputs: neuron projection.

\textbf{Turn 1:}\\
\begin{safebox}
I need to break down this question step by step:

1. Find the disease "Aphasia, Primary Progressive"
2. Find which genes are downregulated in this disease
3. Find which cellular components are associated with those genes
4. Determine which cellular component is involved with the most genes

Let me first use the node_id_retriever to find the ID for "Aphasia, Primary Progressive".

<tool_call>
{"name": "node_id_retriever", "arguments": {"queries": ["Aphasia, Primary Progressive"], "topk": 2}}
</tool_call>

<tool_response>
{'topk': 2, 'results': [[{'node_id': 'D018888', 'node_type': 'Symptom', 'score': 1.0, 'name': 'Aphasia, Primary Progressive'}, {'node_id': 'D057178', 'node_type': 'Symptom', 'score': 0.79, 'name': 'Primary Progressive Nonfluent Aphasia'}]]}
</tool_response>
\end{safebox}

\textbf{Turn 2:}\\
\begin{safebox}
I see that "Aphasia, Primary Progressive" is mapped to a Symptom node with ID `D018888`. However, I need to find genes that are downregulated in a disease, not symptoms.

Looking at the relationship types, I can see there's `DISEASE_DOWNREGULATES_GENE` which should connect a disease to genes that are downregulated. But I need to confirm if this disease is actually a disease node or if I need to find the disease entity.

Let me check if there's a disease node with a similar name. Maybe "Aphasia, Primary Progressive" is actually a symptom, and I need to find the underlying disease. Alternatively, let me first check what relationships exist from this symptom node.

I should verify if the disease is actually a disease node by checking the relationship types. Let me first see what relationships this symptom node has, especially looking for disease-related ones.

<tool_call>
{"name": "code_interpreter", "arguments": {"code": "rows = cypher(\"\"\"\nMATCH (s:Symptom {id: 'D018888'})-[r]->(n:biomedical)\nRETURN type(r) AS rel_type, count(DISTINCT n) AS count\nORDER BY count DESC\n\"\"\")\nprint(rows)"}}
</tool_call>
<tool_response>
[{'rel_type': 'DISEASE_PRESENTS_SYMPTOM', 'count': 2}]
</tool_response>
\end{safebox}

\textbf{Turn 3:}\\
\begin{safebox}
Ah! I see it now. The symptom "Aphasia, Primary Progressive" (D018888) is presented by a disease through the `DISEASE_PRESENTS_SYMPTOM` relationship. So I need to find the disease that presents this symptom.

Let me find the disease node that has this symptom as a presentation.

<tool_call>
{"name": "code_interpreter", "arguments": {"code": "rows = cypher(\"\"\"\nMATCH (s:Symptom {id: 'D018888'})-[r:DISEASE_PRESENTS_SYMPTOM]->(d:Disease)\nRETURN d.id AS disease_id, d.name AS disease_name\n\"\"\")\nprint(rows)"}}
</tool_call>
<tool_response>
[{'disease_id': 'DOID:11949', 'disease_name': 'Creutzfeldt-Jakob disease'}, {'disease_id': 'DOID:10652', 'disease_name': "Alzheimer's disease"}]
</tool_response>
\end{safebox}

\textbf{Turn 4:}\\
\begin{safebox}
Interesting - I see that "Aphasia, Primary Progressive" is associated with both Creutzfeldt-Jakob disease and Alzheimer's disease. Since the question specifically mentions "a disease causing Aphasia, Primary Progressive", I need to find which of these has genes downregulated.

I need to find genes that are downregulated in these diseases. The relationship `DISEASE_DOWNREGULATES_GENE` should connect a disease to genes that are downregulated.

Let me check if there's a `DISEASE_DOWNREGULATES_GENE` relationship from either of these diseases to genes.

<tool_call>
{"name": "code_interpreter", "arguments": {"code": "rows = cypher(\"\"\"\nMATCH (d:Disease {id: 'DOID:11949'})-[r:DISEASE_DOWNREGULATES_GENE]->(g:Gene)\nRETURN count(DISTINCT g) AS gene_count\n\"\"\")\nprint(rows)"}}
</tool_call>
<tool_response>
[{'gene_count': 0}]
</tool_response>
\end{safebox}

\textbf{Turn 5:}\\
\begin{safebox}
Let me try the other disease - Alzheimer's disease (DOID:10652).

<tool_call>
{"name": "code_interpreter", "arguments": {"code": "rows = cypher(\"\"\"\nMATCH (d:Disease {id: 'DOID:10652'})-[r:DISEASE_DOWNREGULATES_GENE]->(g:Gene)\nRETURN count(DISTINCT g) AS gene_count\n\"\"\")\nprint(rows)"}}
</tool_call>
<tool_response>
[{'gene_count': 250}]
</tool_response>
\end{safebox}

\textbf{Turn 6:}\\
\begin{safebox}
Great! So Alzheimer's disease (DOID:10652) has 250 genes that are downregulated. Now I need to find which cellular component is involved with most of these genes.

The relationship between genes and cellular components is `GENE_PARTICIPATES_CELLULAR_COMPONENT`.

Let me find all the cellular components that genes downregulated in Alzheimer's disease are associated with, and count how many genes each component has.

I'll first find all the cellular components that are associated with genes downregulated in Alzheimer's disease.

<tool_call>
{"name": "code_interpreter", "arguments": {"code": "rows = cypher(\"\"\"\nMATCH (d:Disease {id: 'DOID:10652'})-[r:DISEASE_DOWNREGULATES_GENE]->(g:Gene)\nMATCH (g)-[r2:GENE_PARTICIPATES_CELLULAR_COMPONENT]->(cc:Cellular_Component)\nRETURN cc.id AS cc_id, cc.name AS cc_name, count(DISTINCT g) AS gene_count\nORDER BY gene_count DESC\nLIMIT 1\n\"\"\")\nprint(rows)"}}
</tool_call>
<tool_response>
[{'cc_id': 'GO:0043005', 'cc_name': 'neuron projection', 'gene_count': 35}]
</tool_response>
\end{safebox}

\textbf{Turn 7:}\\
\begin{safebox}
Let me get the full list to confirm this is indeed the one with the most genes.

<tool_call>
{"name": "code_interpreter", "arguments": {"code": "rows = cypher(\"\"\"\nMATCH (d:Disease {id: 'DOID:10652'})-[r:DISEASE_DOWNREGULATES_GENE]->(g:Gene)\nMATCH (g)-[r2:GENE_PARTICIPATES_CELLULAR_COMPONENT]->(cc:Cellular_Component)\nRETURN cc.id AS cc_id, cc.name AS cc_name, count(DISTINCT g) AS gene_count\nORDER BY gene_count DESC\n\"\"\")\nprint(rows)"}}
</tool_call>
<tool_response>
[{'cc_id': 'GO:0043005', 'cc_name': 'neuron projection', 'gene_count': 35},
 {'cc_id': 'GO:0045202', 'cc_name': 'synapse', 'gene_count': 30},
 {'cc_id': 'GO:0098796', 'cc_name': 'membrane protein complex', 'gene_count': 27},
 ...]
</tool_response>
\end{safebox}

\textbf{Turn 8:}\\
\begin{safebox}
The result is truncated. Let me get the full result with proper pagination or by limiting to just the top one with the maximum count.

I'll get the exact maximum count by ensuring I get all results properly.

<tool_call>
{"name": "code_interpreter", "arguments": {"code": "rows = cypher(\"\"\"\nMATCH (d:Disease {id: 'DOID:10652'})-[r:DISEASE_DOWNREGULATES_GENE]->(g:Gene)\nMATCH (g)-[r2:GENE_PARTICIPATES_CELLULAR_COMPONENT]->(cc:Cellular_Component)\nRETURN cc.id AS cc_id, cc.name AS cc_name, count(DISTINCT g) AS gene_count\nORDER BY gene_count DESC\n\"\"\")\nprint(rows)"}}
</tool_call>
<tool_response>
[{'cc_id': 'GO:0043005', 'cc_name': 'neuron projection', 'gene_count': 35},
 {'cc_id': 'GO:0045202', 'cc_name': 'synapse', 'gene_count': 30},
 ...]
</tool_response>
\end{safebox}





\textbf{Turn 9:}\\
\begin{safebox}
Perfect! This confirms that the cellular component "neuron projection" is involved with the most genes (35 genes) among those downregulated in Alzheimer's disease, which is the disease causing "Aphasia, Primary Progressive".

So the final answer is "neuron projection".

\answer{neuron projection}
\end{safebox}

\subsubsection{\textbf{Why GraphCoT fails but GraphScout succeeds}}
The failure of GraphCoT mainly stems from a \emph{tool-to-schema mismatch} and the lack of a robust recovery strategy. 
Specifically, GraphCoT assumes that ``Aphasia, Primary Progressive'' is a \texttt{Disease} node and repeatedly queries non-existent neighbor types (e.g., \texttt{Disease-downregulates-Gene}, \texttt{Disease-associates-Gene}). 
Once the tool reports that the neighbor type does not exist, GraphCoT keeps retrying with the same (incorrect) schema assumption, and eventually concludes that the graph contains no relevant information, leading to an incorrect termination.

In contrast, GraphScout exhibits a \emph{two-level advantage} that combines \textbf{reasoning} and \textbf{tool usage}.
On the reasoning level, it explicitly diagnoses the entity-type mismatch (the retrieved node is a \texttt{Symptom}, not a \texttt{Disease}) and formulates a corrective plan: first resolve the underlying disease(s) that present the symptom, then query downregulated genes, and finally aggregate cellular components.
On the tool level, GraphScout leverages executable Cypher queries to (i) validate available relation types from the retrieved node, (ii) traverse the graph to candidate diseases, (iii) verify which candidate disease actually contains \texttt{DISEASE\_DOWNREGULATES\_GENE} edges, and (iv) perform aggregation (\texttt{count(DISTINCT g)}) over \texttt{GENE\_PARTICIPATES\_CELLULAR\_COMPONENT}. 
This synergy between schema-aware reasoning and executable graph operations enables GraphScout to both recover from initial ambiguity and produce a verifiable answer, correctly identifying \emph{neuron projection} as the most frequent cellular component among downregulated genes.

\subsection{Prompt Design for Graph Quizzer}
\label{app:recon_prompt}
Graph Quizzer relies on a carefully structured prompting scheme to guide a strong language model to perform task-oriented yet flexible exploration over an open and programmable graph interface. 
Instead of using rigid question templates, the prompts specify \emph{abstract task objectives}, including question pattern, answer type, and difficulty level, while leaving the concrete exploration strategy to the LLM.
This design encourages diverse graph exploration behaviors while ensuring that the collected evidence can be reliably grounded in the underlying knowledge graph.

\begin{tcolorbox}[title={Base System Prompt for Graph Quizzer}, colback=gray!5, colframe=gray!60]
You are a graph-analysis agent operating on a heterogeneous knowledge graph stored in Neo4j.

Your objective is to perform multi-round, task-oriented exploration to collect graph evidence that supports a specific question pattern, answer type, and difficulty constraint.

You interact with the graph through a programmable Python sandbox with access to a read-only Cypher interface.
Your exploration should be:
\begin{itemize}
    \item \textbf{Task-oriented}: focus on evidence relevant to the current task specification.
    \item \textbf{Incremental}: gather information over multiple rounds rather than exhaustively traversing the graph.
    \item \textbf{Verifiable}: ensure that all collected evidence can be reproduced via explicit Cypher queries.
\end{itemize}

Once sufficient evidence has been collected, you should stop exploration and construct exactly one graph reasoning question together with its answer.
\end{tcolorbox}

\subsubsection{\textbf{Difficulty-Specific Exploration Prompts}} controls the expected depth and structural complexity of exploration, ranging from shallow, local queries to multi-hop, compositional reasoning over heterogeneous graph structures.
It determines the allowed traversal depth, the number of intermediate entities, and whether hybrid reasoning patterns are permitted.\\

\begin{tcolorbox}[title={Difficulty Level: SIMPLE}, colback=bluegray, colframe=bluegrayframe]
\textbf{Design Goal.}
Encourage shallow, interpretable graph exploration that can be resolved with minimal reasoning steps.

\textbf{Exploration Constraints.}
\begin{itemize}
    \item Restrict traversal depth to 1--2 hops.
    \item Prefer a single dominant relationship type.
    \item Avoid compositional or hybrid reasoning patterns.
\end{itemize}

\textbf{Evidence Scope.}
Collected evidence should support direct fact retrieval or simple aggregation, and be sufficient to answer the question without multi-step inference.

\textbf{Typical Use Cases.}
Single-hop entity lookup, attribute retrieval, or basic neighborhood queries.
\end{tcolorbox}

\begin{tcolorbox}[title={Difficulty Level: MEDIUM}, colback=bluegray, colframe=bluegrayframe]
\textbf{Design Goal.}
Promote controlled multi-step reasoning while maintaining clear semantic structure.

\textbf{Exploration Constraints.}
\begin{itemize}
    \item Explore 2--4 hop paths with explicit intermediate nodes.
    \item Involve at least two distinct relationship types or node types.
    \item Allow limited use of filtering, ranking, or property constraints.
\end{itemize}

\textbf{Evidence Scope.}
The collected evidence should form a clear reasoning chain that can be verbalized into a natural-language question.

\textbf{Typical Use Cases.}
Multi-hop entity discovery, constrained aggregation, or relational queries requiring intermediate reasoning.
\end{tcolorbox}

\begin{tcolorbox}[title={Difficulty Level: HARD}, colback=bluegray, colframe=bluegrayframe]
\textbf{Design Goal.}
Encourage non-trivial, compositional graph reasoning that cannot be resolved by local lookup.

\textbf{Exploration Constraints.}
\begin{itemize}
    \item Allow hybrid or combined question patterns.
    \item Explore 3--5 hop heterogeneous paths.
    \item Require traversal through multiple intermediate entities and relation types.
    \item Explicitly avoid staying within a single entity’s immediate neighborhood.
\end{itemize}

\textbf{Evidence Scope.}
Evidence must support multi-step dependency between reasoning stages, ensuring that the final answer depends on earlier intermediate results.

\textbf{Typical Use Cases.}
Recommendation-style reasoning, cross-entity synthesis, and complex relational verification.
\end{tcolorbox}

\subsubsection{\textbf{Query Pattern-Specific Exploration Prompts}} specifies the abstract structural form of a graph question using a (possibly incomplete) triple representation.
Rather than fixing concrete templates, the pattern defines which components are given and which must be discovered through exploration, thereby guiding the direction and scope of evidence collection.
\\
\begin{tcolorbox}[title={Pattern: $\langle h, \_, \_ \rangle$ (Entity-Centric Exploration)}, colback=teal!5, colframe=teal!55]
\textbf{Task Definition.} Only the subject entity $h$ is given, while both the predicate and object are unspecified.

\textbf{Exploration Objective.}
Collect broad but structured information about the subject entity, including:
\begin{itemize}
    \item the node type and key attributes of $h$,
    \item dominant outgoing relation types,
    \item representative neighboring entities and their types,
    \item simple graph statistics.
\end{itemize}

\textbf{Design Constraint.}
Exploration should remain local and avoid aggregating excessive or irrelevant neighborhoods.

\end{tcolorbox}

\begin{tcolorbox}[title={Pattern: $\langle h, r, * \rangle$ (Object-Finding)}, colback=teal!5, colframe=teal!55]
\textbf{Task Definition.} The subject entity $h$ and predicate $r$ are specified, while the object is unknown.

\textbf{Exploration Objective.}
Identify target entities that satisfy the given relation(s), possibly involving:
\begin{itemize}
    \item single-hop or multi-hop predicate chains,
    \item filtering or aggregation constraints,
    \item explicit intermediate nodes as supporting evidence.
\end{itemize}

\textbf{Design Constraint.}
The exploration must yield a reproducible query result and avoid ambiguous answer sets.

\end{tcolorbox}

\begin{tcolorbox}[title={Pattern: $\langle h, *, t \rangle$ (Relationship Discovery)}, colback=teal!5, colframe=teal!55]
\textbf{Task Definition.} Both subject $h$ and object $t$ are given, while the relationship between them is unknown.

\textbf{Exploration Objective.}
Discover and explain how $h$ and $t$ are connected in the graph by:
\begin{itemize}
    \item enumerating valid relation paths,
    \item identifying intermediate entities,
    \item selecting concise and interpretable paths as evidence.
\end{itemize}

\textbf{Design Constraint.}
Preference is given to short or semantically meaningful paths rather than exhaustive enumeration.

\end{tcolorbox}

\begin{tcolorbox}[title={Pattern: $\langle h, r, t \rangle$ (Verification)}, colback=teal!5, colframe=teal!55]
\textbf{Task Definition.} Subject, predicate, and object are all specified.

\textbf{Exploration Objective.}
Verify whether the specified relationship (or constrained relation chain) holds in the graph.

\textbf{Evidence Requirement.}
The final answer must be derived from an explicit verification query (e.g., existence check or count-based test) and supported by at least one concrete witness when the result is positive.

\end{tcolorbox}

\begin{tcolorbox}[title={Hybrid Pattern (Compositional Graph Reasoning)}, colback=teal!5, colframe=teal!55]
\textbf{Task Definition.}
Combine exactly two different question patterns into a single coherent question, where the second reasoning step depends on the result of the first.

\textbf{Exploration Objective.}
Ensure that:
\begin{itemize}
    \item both reasoning steps are necessary to answer the question,
    \item intermediate results from the first step are explicitly reused,
    \item the final question cannot be answered by a single local lookup.
\end{itemize}

\end{tcolorbox}

\subsubsection{\textbf{Answer-Type-Specific Prompt Constraints}} imposes explicit constraints on the form and computability of the final answer, including entity, boolean, numeric, and set-valued outputs.
Each answer type is associated with verifiable evidence requirements (e.g., uniqueness, existence, aggregation), ensuring that generated questions admit deterministic, graph-grounded supervision. \\
\begin{tcolorbox}[title={Answer Type: ENTITY}, colback=violet!5, colframe=violet!55]
\textbf{Objective.}
Produce exactly one uniquely identifiable real-world entity as the answer.

\textbf{Hard Constraints.}
\begin{itemize}
    \item The candidate answer set must be reduced to exactly one entity.
    \item The model must verify uniqueness using an explicit Cypher aggregation.
    \item Internal node identifiers are forbidden; only human-readable names are allowed.
\end{itemize}

\end{tcolorbox}

\begin{tcolorbox}[title={Answer Type: BOOLEAN}, colback=violet!5, colframe=violet!55]
\textbf{Objective.}
Verify whether a specified relationship or constraint holds in the graph.

\textbf{Hard Constraints.}
\begin{itemize}
    \item The boolean value must be derived from an explicit verification query (e.g., existence check or count-based test).
    \item TRUE is returned iff the computed count is strictly greater than zero.
    \item At least one concrete witness must be identified when the result is TRUE.
\end{itemize}

\end{tcolorbox}

\begin{tcolorbox}[title={Answer Type: NUMBER}, colback=violet!5, colframe=violet!55]
\textbf{Objective.}
Return a single numeric value computed from the graph.

\textbf{Hard Constraints.}
\begin{itemize}
    \item The number must be produced by a Cypher aggregation query returning exactly one row.
    \item LIMIT clauses are forbidden in final counting queries.
\end{itemize}

\end{tcolorbox}

\begin{tcolorbox}[title={Answer Type: SET}, colback=violet!5, colframe=violet!55]
\textbf{Objective.}
Return a small, interpretable set of entities.

\textbf{Hard Constraints.}
\begin{itemize}
    \item The total set size must be computed explicitly.
    \item If the set size exceeds a predefined threshold (e.g., 10), additional constraints must be introduced.
    \item The final answer must be a comma-separated list of entity names.
\end{itemize}

\end{tcolorbox}

\subsection{Prompt Design for Graph Solver}

\subsubsection{\textbf{Tool Interface Formatting}}
\label{app:tools}
We adopt the Qwen3 chat template, which automatically injects tool descriptions into the model prompt at runtime. When a non-empty tool set is provided, the template constructs a system-level instruction that appends a dedicated Tools block containing the JSON schemas of all available tools. This block is enclosed by $<tools>…</tools>$ tags and precedes the dialogue history. The same system message also specifies a strict tool invocation protocol, requiring the model to emit function calls as structured JSON objects wrapped by $<tool\_call>…</tool\_call>$, and received by $<tool\_response>…</tool\_response>$. As a result, both tool semantics and calling conventions are explicitly grounded in the prompt, enabling deterministic parsing and execution of tool calls during inference.

\begin{tcolorbox}[title={Code Interpreter Tool Description}, colback=gray!5, colframe=gray!60]
\textbf{tool\_schema:}\\
\hspace*{0.5em}\textbf{type:} "function"\\
\hspace*{0.5em}\textbf{function:}\\
\hspace*{1em}\textbf{name:} "code\_interpreter"\\
\hspace*{1em}\textbf{description:} "Execute Python code for querying and analyzing a heterogeneous graph stored in Neo4j. Inside the executed Python, use the provided synchronous function `cypher(query, params=None, limit=None)` (not a separate tool) to run read-only Cypher queries."\\
\hspace*{1em}\textbf{parameters:}\\
\hspace*{1.5em}\textbf{type:} "object"\\
\hspace*{1.5em}\textbf{properties:}\\
\hspace*{2em}\textbf{code:}\\
\hspace*{2.5em}\textbf{type:} "string"\\
\hspace*{2.5em}\textbf{description:} "Python code"\\
\hspace*{1.5em}\textbf{required:} ["code"]

\end{tcolorbox}

\begin{tcolorbox}[title={Node Retriever Tool Description}, colback=gray!5, colframe=gray!60]
\textbf{tool\_schema:}\\
\hspace*{0.5em}\textbf{type:} "function"\\
\hspace*{0.5em}\textbf{function:}\\
\hspace*{1em}\textbf{name:} "node\_id\_retriever"\\
\hspace*{1em}\textbf{description:} "Retrieve top nodes ids from a heterogeneous graph by semantic similarity (batch). You can use this tool to get target nodes ids first and then continue next steps."\\
\hspace*{1em}\textbf{parameters:}\\
\hspace*{1.5em}\textbf{type:} "object"\\
\hspace*{1.5em}\textbf{properties:}\\
\hspace*{2em}\textbf{queries:}\\
\hspace*{2.5em}\textbf{type:} "array"\\
\hspace*{2.5em}\textbf{description:} "A list of query strings. Such as ['Nausea', 'MTUS1']"\\
\hspace*{2.5em}\textbf{items:}\\
\hspace*{3em}\textbf{type:} "string"\\
\hspace*{2em}\textbf{topk:}\\
\hspace*{2.5em}\textbf{type:} "integer"\\
\hspace*{2.5em}\textbf{description:} "Number of retrieved nodes per query. Default is 2."\\
\hspace*{2.5em}\textbf{default:} 2\\
\hspace*{1.5em}\textbf{required:} ["queries"]

\end{tcolorbox}

\subsubsection{\textbf{Formatting of Graph Descriptions}} To ensure clarity and reproducibility, we explicitly standardize the formatting of graph descriptions and code examples in the Graph Solver.

\begin{tcolorbox}[title={System Prompt}, colback=gray!5, colframe=gray!60]
You are given a $<domain\_label>$ heterogeneous knowledge graph stored in Neo4j.\\
\hspace*{0.5em}- Backend: Neo4j (Cypher)\\
\hspace*{0.5em}- Label scope: :$<domain\_label>$\\
\hspace*{0.5em}- Node properties:\\
\hspace*{1em}- The dataset has these node\_types: $<node\_type\_list>$\\
\hspace*{1em}- each node\_type has its own set of type-specific properties: $<properties\_of\_each\_type>$\\
\hspace*{1em}- There are several types of edges (relationship type) in this graph: $<edge\_type\_list>$\\
\hspace*{0.5em}- How to query in code: $<code\_examples>$\\
- Output rule: The very last line of your response must be exactly in the form \texttt{\textbackslash answer\{FINAL\_ANSWER\}}. Do not include any extra text on that last line.
\end{tcolorbox}


\end{document}